\pdfoutput=1

\documentclass[11pt,table,dvipsnames]{article}
\usepackage[final]{acl}

\usepackage{times}
\usepackage{latexsym}
\usepackage{makecell}
\usepackage{adjustbox}
\usepackage{array}

\usepackage{booktabs,longtable}
\usepackage[T1]{fontenc}

\usepackage[utf8]{inputenc}

\usepackage{microtype}

\usepackage{inconsolata}

\usepackage{graphicx}

\usepackage{amsmath,amsfonts}
\usepackage{algorithmic}
\usepackage{array}
\usepackage{textcomp}
\usepackage{stfloats}
\usepackage{url}
\usepackage{verbatim}
\usepackage{multirow}
\usepackage{hyperref}
\usepackage{arydshln}
\usepackage{comment}
\usepackage{booktabs}
\usepackage{xcolor}
\usepackage{lipsum}
\usepackage{subcaption}
\usepackage{float}
\usepackage{changepage}
\usepackage{enumitem}    
\usepackage{amssymb}     
\usepackage{pifont}      

\definecolor{lightblue}{RGB}{91, 155, 213}
\definecolor{medblue}{RGB}{68, 114, 196}
\definecolor{darkblue}{RGB}{47, 85, 151}
\definecolor{lightgray}{RGB}{166, 166, 166} 
\definecolor{medgray}{RGB}{128, 128, 128} 
\definecolor{darkgray}{RGB}{89, 89, 89}
\definecolor{lightred}{RGB}{255, 102, 102}    
\definecolor{medred}{RGB}{204, 0, 0}      
\definecolor{darkred}{RGB}{139, 0, 0}

\title{ Entailed Opinion Matters: Improving the Fact-Checking Performance of Language Models by Relying on their Entailment Ability}

\author{
Gaurav Kumar, Ayush Garg, Debajyoti Mazumder, Aditya Kishore, Babu kumar, Jasabanta Patro \\
   Department of Data Science and Engineering \\
  Indian Institute of Science Education and Research, Bhopal, India \\
  \texttt{\{gaurav22, ayushg24, debajyoti22, adityak21, babu21, jpatro\}@iiserb.ac.in}\\
}

\begin{document}
\maketitle
\begin{abstract}
Automated fact-checking has been a challenging task for the research community. Prior work has explored various strategies, such as end-to-end training, retrieval-augmented generation, and prompt engineering, to build robust fact-checking systems. However, their accuracy has not been high enough for real-world deployment. We, on the other hand, propose a new learning paradigm, where evidence classification and entailed justifications made by generative language models (GLMs) are used to train encoder-only language models (ELMs). We conducted a rigorous set of experiments, comparing our approach with recent works along with various prompting and fine-tuning strategies. Additionally, we performed ablation studies, error analysis, quality analysis of model explanations, and a domain generalisation study to provide a comprehensive understanding of our approach.
\end{abstract}


\section{Introduction:}

\begin{figure}[t!]
\centering
\includegraphics[width=0.9\columnwidth]{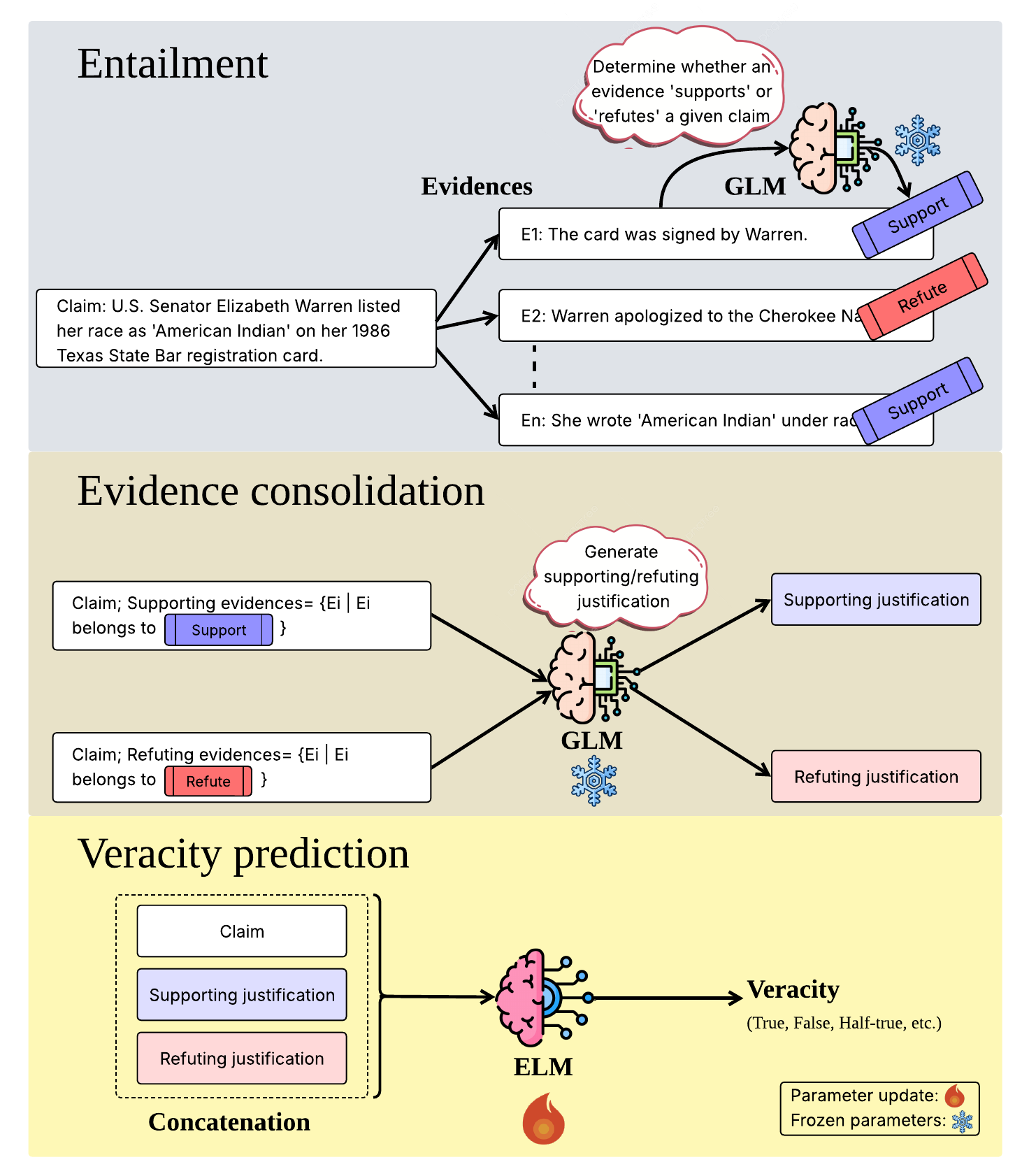}
\caption{A schematic diagram explaining our learning paradigm, which breaks down into three stages. (a) \textit{Entailment}: Given a claim and its corresponding evidences, a GLM first distinguishes evidences into supporting/ refuting evidence via entailment. (b) \textit{Evidence consolidation}: Then the same GLM consolidates these two groups into concise supporting and refuting justification. (c) \textit{Veracity prediction}: Using the claim and both justifications, an ELM is trained to predict veracity.
}
\label{fig:motivation_ex}
 \vspace{-2mm}
\end{figure}

The spread of misinformation on the internet has become a pressing social issue. Its consequences have manifested across social, political, and commercial domains \cite{mozur2018genocide, allcott2017social}. Manually verifying facts and detecting misinformation is a slow and costly process. To overcome this, the research community has proposed several approaches for automated fact-checking. Their interest can be gauged by the fact that more than 1,200 research articles and 50 datasets spanning different languages and modalities have been published on this topic \cite{alnabhan2024fake, guo2022survey}. Most of them use deep neural architectures to learn latent features for a given input. Researchers, in this case, use attention maps to assess the model's decision-making process. Recently, with the emergence of generative language models (GLMs), researchers have begun exploring prompt engineering approaches. Zero-shot prompting, few-shot prompting, and CoT-based prompting \cite{qiao2023reasoning} have been among the front-runners. However, the accuracy of existing systems is not high enough to deploy them in the real world. The primary reason behind this is that fact verification is a complex process. For a given claim, language models have to parse through multiple pieces of evidence, which often contradict each other. The contradictory evidence confuses the language models, leading to poor performance.

Prompt-engineering approaches, on the other hand, have achieved human-level performance when used in conjunction with a large generative language model (with over a trillion parameters) \cite{qiao2023reasoning}. However, their deployment is costly, as large GLMs usually remain behind a paywall. Using prompting approaches on smaller GLMs such as Mistral-7B \cite{jiang2023mistral7b}, Llama-7B \cite{meta2024llama3}, etc., may seem to be a solution, but (i) their performance is lower than training-based approaches, and (ii) the model outcome often suffers from bias and hallucination. To overcome this, we propose a simple yet effective strategy, formulating a new training paradigm for fact-checking. Here, we classify evidence and generate entailed justifications using GLMs, and finetune ELMs for veracity predictions based on these justifications. Particularly, we follow a three-step process as illustrated in Figure \ref{fig:motivation_ex}. We conducted a variety of training (TBE)- and inference (IBE)-based experiments to evaluate the effectiveness of our approach across different input scenarios. All of our experiments were done around three research questions. They are,

\begin{itemize}
    \item \textbf{R1:} \textit{"How well do the language models perform when only claim and raw evidence sentences are available?"}
    \item \textbf{R2:} \textit{"Does prompted claim-evidence understanding improve the performance of the language models?"}
    \item \textbf{R3:} \textit{"Can prompted entailed justifications improve the claim veracity prediction?"}
\end{itemize}

Some of our key contributions and observations in this work can be summarised as follows:

\begin{itemize}
    \item We conducted three training-based (TBEs) and four inference-based (IBEs) experiments along the line of R1, R2, and R3 for two monolingual English datasets, i.e., LIAR-RAW and RAW-FC. For RAW-FC, training with raw evidence sentences (TBE-1) and overall claim-evidence understanding (TBE-2) registered an improvement of up to \textbf{8.20\%} and \textbf{16.39\%} in macro-F1 ($MF1$) over the baselines. Training with the entailed justifications (TBE-3), on the other hand, outperformed the baselines by a significant margin (up to \textbf{28.57\%} and \textbf{44.26\%} for LIAR-RAW and RAW-FC, respectively).

    \item We performed an ablation study by removing individual supporting and refuting entailed justifications. We found that both of the entailed justifications were important, as their removal led to a decline in $MF$ performance.
    
    \item We considered the entailed justifications generated by GLMs as model explanations and evaluated them using two strategies: (i) checking lexical overlap and semantic matching, and (ii) conducting subjective evaluation by the GLMs themselves. Notably, explanations generated by Llama received the highest subjective scores, correlating with the highest veracity prediction performance. We performed a thorough linguistic analysis of the explanations by looking into the model attention scores. We found that ELMs assigned higher attention scores to appropriate keywords and factual elements in samples with extreme labels (e.g., true or false). However, for borderline samples (e.g., half-true labels), attention scores seemed scattered. These studies were kept in the appendix due to space constraints.  
    
    \item We extended our evaluation (TBE-3) to multilingual (X-Fact and Ru22fact) and multimodal (Factify-2, MOCHEG, and VERITE) datasets to check its domain generalisability. Our approach resulted in performance improvements (in $MF1$) of up to \textbf{19.70\%} in multilingual and \textbf{54.10\%} in multimodal domains.
\end{itemize}

\section{Related Works:}
\label{section:related_works}

Automated fact-checking has been extensively studied in the literature, with prior work spanning diverse datasets, modeling approaches, and task settings \cite{guo2022survey, vladika-matthes-2023-scientific}. While most fact-checking research has focused on text-only, English-language claims, recent works have expanded the literature to multimodal and multilingual settings \cite{akhtar2023multimodal, tufchi2023comprehensive}.
\vspace{-0.5em}

\paragraph{Overall approaches:} Early approaches relied heavily on manual feature engineering and retrieval from external knowledge sources \cite{guo2022survey, zeng2021automated}. With the rise of deep learning, encoder-only language models (ELMs) such as BERT-like architectures became a dominant approach through fine-tuning \cite{lee-etal-2020-language}. In recent years, larger generative language models (GLMs) have shown dominance in this field due to their vast pretraining knowledge \cite{vykopal2024generative}. While effective, language models often struggle with bias and hallucination when evidence is noisy or contradictory \cite{10.1145/3703155}.

\paragraph{Handling evidence:} To mitigate these issues, recent works have introduced evidence discrimination, aggregation, and conflict resolution methods. State-of-the-art pipelines have explored diverse strategies for handling evidence in fact-checking. For instance, \citet{yue2024retrieval} used retrieved evidence to generate supporting and opposing arguments and applied few-shot prompting for claim verification. However, they did not explicitly divide evidence sets that may contain contradictory statements. Similarly, L-Defense \cite{wang2024explainable} attempted to filter or weight evidence before prediction. However, the authors still relied on complex attention mechanisms (or learned relevance scores), which may fail when many pieces of evidence conflict with each other. Apart from that, \citet{wang2024explainable} relied on attention scores that may discard useful information by focusing only on top-k evidence. In contrast, we argue that GLMs, with their strong understanding of language \cite{10.1145/3701716.3717815}, can explicitly analyze how each piece of evidence supports or refutes a claim. Their ability to process long contexts allows them to consider the entire evidence set rather than a small subset.

\vspace{-0.5em}

\paragraph{Entailment:} Entailment is the process of determining whether a hypothesis logically follows from a premise and has long been a key component in fact-checking pipelines \cite{luken-etal-2018-qed, MARTIN2022109265}. However, prior works have primarily used entailment for claim-evidence matching or similarity scoring rather than as an explicit evidence discrimination step.

\vspace{-0.5em}

\paragraph{Research Gap:} Our work aimed to bridge this gap by explicitly incorporating entailment as the initial step, leveraging GLMs to classify each evidence piece as supporting or refuting the claim, and then consolidating these groups into concise justifications that served as input for training ELMs to predict veracity. To validate this hypothesis, we compared it against various training- and inference-based methods, as well as state-of-the-art baselines. To the best of our knowledge, this constitutes a new learning paradigm that is simple and effective across modalities. None of the prior works have leveraged GLM-based entailment as a primary signal to discriminate between supporting and refuting evidence for downstream veracity prediction.
\section{Dataset Details:}

In this section, we reported the details of the datasets considered in our study. We considered the \textbf{LIAR-RAW} and \textbf{RAW-FC} datasets provided by \citet{yang-etal-2022-coarse} to compare the performance of all TBEs and IBEs. Additionally, we considered two multilingual datasets, i.e., X-FACT \cite{gupta2021x} and RU22Fact \cite{zeng2024ru22fact}, and three multimodal datasets, i.e., Factify-2 \cite{DBLP:conf/defactify/SuryavardanMPCR23}, MOCHEG \cite{10.1145/3539618.3591879}, and VERITE \cite{papadopoulos2024verite}, to check the domain generalisability of our learning paradigm. The basic statistical details of the datasets are reported in Table \ref{tab:dataset_stats}. More details on the datasets are provided in Appendix \ref{Additional_dataset_details}.



\begin{table}[t]
\centering
\small
\setlength{\tabcolsep}{9pt}      
\renewcommand{\arraystretch}{1.15}

\resizebox{\columnwidth}{!}{%
\begin{tabular}{l@{\hspace{20pt}}c c r r r r}
\toprule
\textbf{Datasets} & \textbf{No. of Classes} & \textbf{Language} &
\textbf{Train} & \textbf{Val} & \textbf{Test} & \textbf{Total} \\
\midrule
\multicolumn{7}{l}{\textsc{Monolingual}} \\
~~~-LIAR-RAW & 6 & English       & 10,065 & 1274 & 1251 & 12,590 \\
~~~-RAW-FC   & 3 & English       & 1,612 & 200 & 200 & 2,012 \\
\midrule

\multicolumn{7}{l}{\textsc{Multimodal}} \\
~~~-Factify-2 & 3 & English       & 27,500 & 7,500 & 7,500 & 42,500 \\
~~~-MOCHEG   & 3 & English       & 11,669 & 1,490 & 2,442 & 15,601 \\
~~~-VERITE   & 3 & English       & 811    & 101   & 102   & 1,014  \\
\midrule

\multicolumn{7}{l}{\textsc{Multilingual}} \\
~~~-X-FACT    & 7 & 25 languages  & 19,079 & 2,535 & 9,575 & 31,189 \\
~~~-RU22Fact  & 3 & 4 languages   & 11,217 & 1,600 & 3,216 & 16,033 \\
\bottomrule
\end{tabular}%
}

\caption{Details of monolingual, multilingual and multimodal datasets considered in our work. While monolingual datasets are used to evaluate the performance of TBEs and IBEs, others are used in the domain generalisation study.}
\label{tab:dataset_stats}
\end{table}

\vspace{-6pt}
\section{Experiments:} 
As mentioned previously, we conducted three training-based (\textbf{TBEs}) and four inference-based (\textbf{IBEs}) experiments. The schematic diagram showing all TBEs and IBEs is illustrated in Figure \ref{fig:TBE_overview}. In TBEs, we finetuned (i) ELMs like RoBERTa \cite{liu2019roberta} and XLNet \cite{yang2019xlnet}, and (ii) GLMs like Mistral \cite{jiang2023mistral7b}, Llama \cite{meta2024llama3}, Gemma \cite{team2024gemma}, Qwen \cite{yang2024qwen25}, and Falcon \cite{almazrouei2023falcon} with LoRA \cite{hu2022lora} and LoRA+ \cite{hayou2024lora+} adapters. In IBEs, on the other hand, we prompted the GLMs to predict the veracity. For domain generalisation experiments, we considered appropriate ELMs and GLMs that could process multilingual and multimodal input (see Section \ref{sec:domain_gen}). The maximum supported input length and the versions of different ELMs and GLMs that we considered are reported in Table \ref{tab:LLM_VLLM_Details} (in the appendix). Note that unlike ELMs, we could not directly fine-tune GLMs (without adapters) due to their large parameter size and our computational constraints. The details of individual experiments and the current state-of-the-art baselines are reported in the subsequent subsections. The details of the experimental set-ups, including hyperparameters, the choice of evaluation metrics, and the constraints we faced for model reproducibility, are reported in Appendix \ref{sec:exp_setup}.


\subsection{Training Based Experiments (\textbf{TBEs}):}
\label{sec:tbe}

\subsubsection{TBE-1: Training based on raw-evidences:}

Our first training-based experiment was based on R1, in which we fine-tuned the GLMs (using LoRA and LoRA+) using claims and raw evidence sentences. If any sample lacked an associated evidence sentence, we provided only the claim as input. We restricted ourselves from using ELMs like RoBERTa \cite{liu2019roberta} and XLNet \cite{yang2019xlnet}, as the length of many input instances exceeded their maximum supported input length (Table \ref{tab:LLM_VLLM_Details}). Past work \cite{cheung2023factllama} demonstrated the effectiveness of fine-tuning Llama with LoRA adapters using evidence pieces from web search, whereas in our case, we used the gold evidence sentences given in the dataset. To the best of our knowledge, we were the first to fine-tune GLMs using raw evidence sentences. Our approach is illustrated in sub-figure (a) of Figure \ref{fig:TBE_overview}.

\subsubsection{TBE-2: Training based on overall understanding:}
Our second experiment in this line was based on R2. To conduct it, first, we prompted the five considered GLMs to generate their \textit{understanding} of a given claim and its evidence pieces. For samples without an associated evidence sentence, GLMs generated their \textit{understandings} solely from the claim. With that, we fine-tuned the ELMs and GLMs (with adapters) to produce the claim veracity. The detailed experimental process is illustrated in sub-figure (b) of Figure \ref{fig:TBE_overview}. The prompts used are reported at ID \#3 in Table \ref{tab:prompts} in the appendix.

\subsubsection{TBE-3: Training based on entailment understanding:}
\label{sec:tbe3}

\begin{figure*}[t]
    \centering
    \includegraphics[width=0.9 \textwidth]{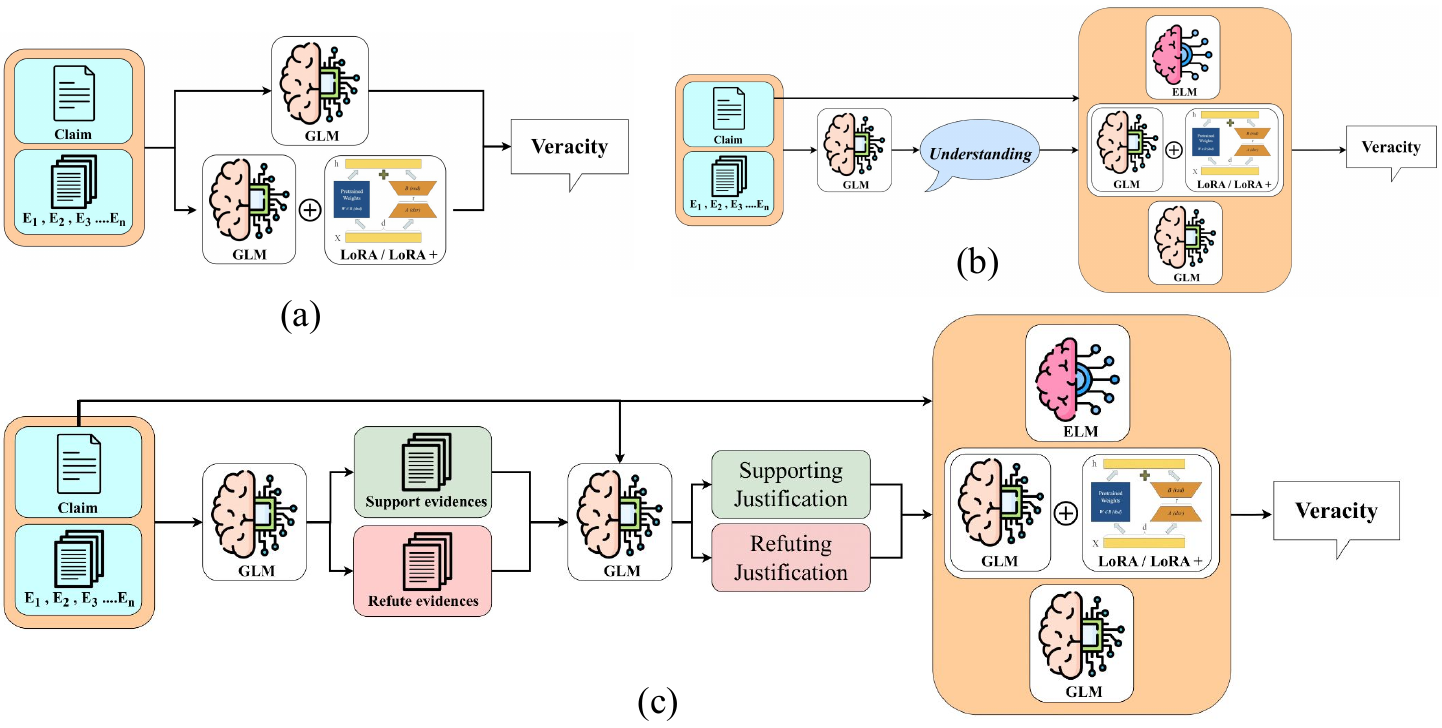}
    \caption{ Illustration of steps we followed in different experiments. Sub-figure (a) presents the case where only raw evidence sentences and claims are given as input (\textbf{R1}). This approach is used in \textbf{TBE-1} and \textbf{IBE-1}. Sub-figure (b) shows the overall process of \textbf{TBE-2}, \textbf{IBE-2} and \textbf{IBE-3} based on \textbf{R2}. Sub-figure (c) illustrates the overall experimental process of \textbf{TBE-3} and \textbf{IBE-4} based on \textbf{R3}.}
    \label{fig:TBE_overview}
\end{figure*}

Our final training-based experiment was based on R3. To conduct it, we followed a three-step approach. In the first step, we prompted the considered GLMs to classify whether the associated evidence sentences were “supporting” or “refuting” a given claim. In the second step, we prompted the language models to generate supporting and refuting \textit{justifications} based on the classified evidence sentences. For cases where claims lacked supporting or refuting evidence sentences, GLMs generated justifications based on their embedded knowledge. Finally, based on the claims and the generated justifications, we fine-tuned (i) the considered ELMs and (ii) adapters with the GLMs to generate the veracity. The detailed approach is illustrated in sub-figure (c) of Figure \ref{fig:TBE_overview}. Some of the prompt samples we used at each step are reported in Table \ref{tab:prompts} (ID 9, 10, and 11) in the appendix.

\subsection{Inference Based Experiments (\textbf{IBEs}):}
\label{sec:ibe}
We conducted four inference-based experiments (IBEs) based on R1, R2, and R3. These included: (i) \textit{zero-shot prompting} (IBE-1), where we prompted the GLMs to predict the claim veracity directly from the claims and associated evidence sentences, unlike prior studies \cite{zhang2023towards, cheung2023factllama} that either used only the claims or relied on web-retrieved evidence; (ii) \textit{zero-shot prompting with overall understanding} (IBE-2), where first we prompted the GLMs to generate claim-evidence understanding and then used it to predict the veracity, a two-step prompting strategy that was not explored previously; (iii) \textit{CoT prompting with overall understanding} (IBE-3), which extended IBE-2 by explicitly instructing GLMs to produce step-by-step reasoning, differing from prior CoT approaches \cite{zhang2023towards} that depended on external evidence retrieval. In contrast, our method leveraged GLM-generated claim-evidence understanding, thereby avoiding dependency on external retrieval; and (iv) \textit{prompting based on entailment} (IBE-4), where we prompted GLMs to predict veracity using entailed justifications, marking the first attempt to classify each evidence sentence as supporting or refuting a claim and then generate entailment-based justifications for fact verification, unlike prior entailment works \cite{10.1145/3589335.3651504,10.1145/3534678.3539205}. For claims without associated evidence, only the claim text was used as input. Representative prompt samples for each step are listed in Table \ref{tab:prompts} (IDs 1–12) in the appendix.

\subsection{Baselines:} 
\label{section:baselines}

In this section, we reported the previously proposed best-performing models as the baselines. Particularly, we compared our models with the performances of HiSS \cite{zhang2023towards}, FactLLaMa \cite{cheung2023factllama}, RAFTS \cite{yue2024retrieval}, and L-Defense \cite{wang2024explainable}. Out of them, HiSS \cite{zhang2023towards} and FactLLaMa \cite{cheung2023factllama} retrieved evidence from external sources, while RAFTS \cite{yue2024retrieval} employed a coarse-to-fine retrieval technique to extract evidence directly from the dataset. In contrast, L-Defense \cite{wang2024explainable} used relevant evidence without additional retrieval. The previously reported performance of these models on the LIAR-RAW and RAW-FC datasets is presented in Table \ref{tab:liar_rawfc_performance}. We also considered appropriate multilingual and multimodal baselines for the domain generalisation study. Their details are reported in Section \ref{sec:domain_gen}.

\begin{table}[t]
\centering
\footnotesize 
\begin{adjustbox}{max width=0.40\textwidth}
\begin{tabular}{l|ccc|ccc}
\hline
\multirow{2}{*}{\textbf{Method}} & \multicolumn{3}{c|}{\textbf{LIAR-RAW}} & \multicolumn{3}{c}{\textbf{RAWFC}} \\
 \cmidrule(lr){2-4} \cmidrule(lr){5-7}
& \textbf{MP} & \textbf{MR} & \textbf{MF1} & \textbf{MP} & \textbf{MR} & \textbf{MF1} \\
\hline
HiSS       & 0.46 & 0.31 & 0.37 & 0.53 & 0.54 & 0.53 \\
FactLLaMA  & 0.32 & 0.32 & 0.30 & 0.56 & 0.55 & 0.55 \\
RAFTS      & \textbf{0.47} & \textbf{0.37} & \textbf{0.42} & \textbf{0.62} & 0.52 & 0.57 \\
L-Defense      \\
~~~-\textit{ChatGPT} & 0.30 & 0.32 & 0.30 & 0.61 & \textbf{0.61} & \textbf{0.61} \\
~~~-\textit{Llama2}  & 0.31 & 0.31 & 0.31 & 0.61 & 0.60 & 0.60 \\
& (0.29\textsuperscript{\dag})  & (0.29\textsuperscript{\dag})  & (0.29\textsuperscript{\dag})  & (0.56\textsuperscript{\dag})  &  (0.56\textsuperscript{\dag})  & (0.56\textsuperscript{\dag}) \\
\hline
\end{tabular}
\end{adjustbox}
\caption{Performance of the considered baselines for LIAR-RAW and RAWFC datasets. 
Notation: our reproduced results are marked as ‘\dag’.}
\label{tab:liar_rawfc_performance}
\vspace{-2mm}
\end{table}

\section{Results and Discussion:}

In this section, we reported the results of our experiments.

\begin{table}[t]
\centering
\footnotesize
\setlength{\tabcolsep}{1pt}
\renewcommand{\arraystretch}{1}
\resizebox{0.8\columnwidth}{!}{%
\begin{tabular}{llccccc}
\hline
\textbf{Dataset} & \textbf{Method} & \textbf{Mistral} & \textbf{Llama} & \textbf{Gemma} & \textbf{Qwen} & \textbf{Falcon} \\
\hline
\multirow{2}{*}{LIAR-RAW}
    & LoRA 
        & \shortstack{\colorbox{white}{0.27}}
        & \shortstack{\colorbox[HTML]{d2e7d6}{\bfseries 0.30}}
        & \shortstack{\colorbox{white}{0.23}}
        & \shortstack{\colorbox[HTML]{c0e7f6}{\underline{0.29}}}
        & \shortstack{\colorbox{white}{0.26}} \\
    & LoRA+ 
        & \shortstack{\colorbox{white}{0.25}}
        & \shortstack{\colorbox[HTML]{c0e7f6}{\underline{0.29}}}
        & \shortstack{\colorbox{white}{0.22}}
        & \shortstack{\colorbox[HTML]{c0e7f6}{\underline{0.29}}}
        & \shortstack{\colorbox[HTML]{c0e7f6}{\underline{0.29}}} \\
\hline
\multirow{2}{*}{RAW-FC}
    & LoRA 
        & \shortstack{\colorbox[HTML]{c0e7f6}{\underline{0.65}}}
        & \shortstack{\colorbox[HTML]{c0e7f6}{\underline{0.65}}}
        & \shortstack{\colorbox{white}{0.57}}
        & \shortstack{\colorbox[HTML]{d2e7d6}{\bfseries 0.66}}
        & \shortstack{\colorbox{white}{0.54}} \\
    & LoRA+ 
        & \shortstack{\colorbox{white}{0.55}}
        & \shortstack{\colorbox[HTML]{c0e7f6}{\underline{0.65}}}
        & \shortstack{\colorbox{white}{0.57}}
        & \shortstack{\colorbox[HTML]{c0e7f6}{\underline{0.65}}}
        & \shortstack{\colorbox{white}{0.63}} \\
\hline
\end{tabular}}
\caption{Performance of models under \textbf{TBE-1} for LIAR-RAW and RAW-FC datasets in terms of $MF1$. Here, we have reported the average values of three random seeds. \colorbox[HTML]{d2e7d6}{\textbf{Green}} and \colorbox[HTML]{c0e7f6}{\underline{Blue}} indicate best and second-best scores, respectively.}
\label{table:tbe1}
\vspace{-2mm}
\end{table}

\subsection{Observations for veracity prediction:}

We reported the $MF1$ of various models under TBE-1, TBE-2, TBE-3, and IBEs in Table \ref{table:tbe1}, Table \ref{table:tbe2_tbe3}, and Table \ref{table:prompting_ibe}, respectively. Our observations are as follows:

\begin{itemize}
    \item For LIAR-RAW, none of the macro-F1 scores reported by \textbf{IBE} models surpassed the best baseline performance. In contrast, for RAW-FC, Llama achieved the highest performance in IBE-2, surpassing the highest baseline performance.

    \item In \textbf{TBE-1}, we observed a clear contrast between the two datasets. For LIAR-RAW, performance remained low and clustered ($MF1$: 0.22--0.30), and neither LoRA nor LoRA+ provided a reliable advantage. In contrast, RAW-FC showed consistent improvements over the best baseline ($MF1$: 0.61). The best result was achieved by Qwen+LoRA ($MF1$: 0.66, $\sim\textbf{8.20}\% \uparrow$), while several configurations reached ($MF1$: 0.65). Overall, the relative improvement on RAW-FC spanned $\sim\textbf{3.28}\%$ to \textbf{8.20}\% over the baseline.
 
    \item In \textbf{TBE-2}, for LIAR-RAW, no model surpassed the best $MF1$ reported by the baselines. However, for RAW-FC, we observed that many models, such as (i) XLNet fine-tuned on Llama understandings ($\sim\textbf{1.64}\% \uparrow$), and (ii) Llama trained with LoRA+ based on Llama understandings ($\sim \textbf{16.39}\% \uparrow$), outperformed the best reported baseline $MF1$.

    \item \textbf{TBE-3} yielded consistent improvements over the baselines on both datasets. For LIAR-RAW, several fine-tuned ELM configurations surpassed the best baseline ($MF1$: 0.42), with relative gains ranging from \(\sim\textbf{4.76}\%\uparrow\) for RoBERTa and XLNet fine-tuned with Falcon-based entailed justifications to \(\sim\textbf{28.57}\%\uparrow\) for XLNet fine-tuned with Llama-based entailed justifications. Additionally, Llama trained with Llama-generated justifications using the LoRA+ adapter also exceeded the baseline by \textbf{16.67}\%. For RAW-FC, the gains were larger, and multiple fine-tuned RoBERTa/XLNet variants exceeded the best baseline ($MF1$: 0.61), achieving improvements from \(\sim\textbf{4.91}\%\uparrow\) to \(\sim\textbf{44.26}\%\uparrow\). The best performance was obtained by RoBERTa fine-tuned with Llama-based entailed justifications ($MF1$: 0.88, \(\sim\textbf{44.26}\%\uparrow\)), closely followed by XLNet fine-tuned with Llama-based entailed justifications ($MF1$: 0.87, \(\sim\textbf{42.62}\%\uparrow\)). Overall, TBE-3 delivered the strongest and most consistent improvements, indicating that entailed justifications served as an effective supervision signal for veracity prediction.
\end{itemize}

\begin{table}[t]
\centering
\footnotesize 
\setlength{\tabcolsep}{4pt}   
\renewcommand{\arraystretch}{1.1}
\resizebox{0.8\columnwidth}{!}{%
\begin{tabular}{lcccc}
\hline
Dataset ($\xrightarrow{}$) & \multicolumn{2}{c}{\textbf{LIAR-RAW}} & \multicolumn{2}{c}{\textbf{RAW-FC}} \\
\cmidrule(lr){2-3} \cmidrule(lr){4-5}
Method ($\downarrow$) & TBE-2 & TBE-3 & TBE-2 & TBE-3 \\
\hline
\rule{0pt}{3ex}\textsc{Fine-tuning} & & & & \\
~~~-\textit{RoBERTa-L\textsubscript{Mistral}} & 0.26 & 0.47$^{***}$ & 0.50 & 0.83$^{***}$ \\
~~~-\textit{RoBERTa-L\textsubscript{Llama}}   & 0.25 & \colorbox[HTML]{c0e7f6}{\underline{0.52$^{***}$}} & 0.49 & \bf\colorbox[HTML]{d2e7d6}{0.88$^{***}$} \\
~~~-\textit{RoBERTa-L\textsubscript{Gemma}}   & 0.27 & 0.48$^{***}$ & 0.50 & 0.49 \\
~~~-\textit{RoBERTa-L\textsubscript{Qwen}}    & 0.28 & 0.46$^{***}$ & 0.48 & 0.71$^{***}$ \\
~~~-\textit{RoBERTa-L\textsubscript{Falcon}}  & 0.27 & 0.44$^{***}$ & 0.48 & 0.64$^{**}$ \\
~~~-\textit{XLNet-L\textsubscript{Mistral}}   & 0.28 & 0.47$^{***}$ & 0.61 & 0.82$^{***}$ \\
~~~-\textit{XLNet-L\textsubscript{Llama}}     & 0.29 & \bf\colorbox[HTML]{d2e7d6}{0.54$^{***}$} & \colorbox[HTML]{c0e7f6}{\underline{0.62}} & \colorbox[HTML]{c0e7f6}{\underline{0.87$^{***}$}} \\
~~~-\textit{XLNet-L\textsubscript{Gemma}}     & 0.25 & 0.42$^{***}$ & 0.50 & 0.46 \\
~~~-\textit{XLNet-L\textsubscript{Qwen}}      & 0.28 & 0.48$^{***}$ & 0.58 & 0.70$^{***}$ \\
~~~-\textit{XLNet-L\textsubscript{Falcon}}    & 0.24 & 0.44$^{***}$ & 0.60 & 0.74$^{***}$ \\
\hline
\rule{0pt}{3ex}\textsc{LoRA} & & & & \\
~~~-\textit{Mistral} & \bf\colorbox[HTML]{d2e7d6}{0.32} & 0.29 & 0.60 & 0.58 \\
~~~-\textit{Llama}   & 0.25 & 0.30 & 0.46 & 0.61$^{*}$ \\
~~~-\textit{Gemma}   & 0.18 & 0.20 & 0.40 & 0.30 \\
~~~-\textit{Qwen}    & 0.21 & 0.32 & 0.53 & 0.46 \\
~~~-\textit{Falcon}  & 0.23 & 0.21 & 0.53 & 0.41 \\
\rule{0pt}{3ex}\textsc{LoRA+} & & & & \\
~~~-\textit{Mistral} & \colorbox[HTML]{c0e7f6}{\underline{0.30}} & 0.27 & 0.60 & 0.46\\
~~~-\textit{Llama}   & 0.23 & 0.49$^{***}$ & \bf\colorbox[HTML]{d2e7d6}{0.71} & 0.82$^{***}$ \\
~~~-\textit{Gemma}   & 0.20 & 0.29 & 0.51 & 0.46 \\
~~~-\textit{Qwen}    & 0.24 & 0.29 & 0.43 & 0.47 \\
~~~-\textit{Falcon}  & 0.28 & 0.36$^{**}$ & 0.53 & 0.50 \\
\hline
\end{tabular}}
\caption{Performance of claim veracity prediction using gold evidences in terms of macro F1 (MF). We report mean values over three random seeds. \colorbox[HTML]{d2e7d6}{\textbf{Green}} and \colorbox[HTML]{c0e7f6}{\underline{Blue}} denote the best and second-best results, respectively. `*' indicates statistically significant improvements, $^{*}: p<0.05$, $^{**}: p<0.01$, and $^{***}: p<0.001$.}
\label{table:tbe2_tbe3}
\vspace{-2mm}
\end{table}

\begin{table}[t]
\centering
\footnotesize
\setlength{\tabcolsep}{2pt}
\renewcommand{\arraystretch}{1.1}
\resizebox{0.75\columnwidth}{!}{%
\begin{tabular}{l|cccc|cccc}
\hline
Dataset ($\rightarrow$) & \multicolumn{4}{c|}{\textbf{LIAR-RAW}} & \multicolumn{4}{c}{\textbf{RAW-FC}} \\
\cline{2-9}
Method ($\downarrow$) & IBE-1 & IBE-2 & IBE-3 & IBE-4 & IBE-1 & IBE-2 & IBE-3 & IBE-4 \\
\hline
\textsc{Prompting} & & & & & & & & \\
~~~-\textit{Mistral} 
& \bf\colorbox[HTML]{d2e7d6}{0.22} 
& \colorbox[HTML]{c0e7f6}{\underline{0.20}} 
& \bf\colorbox[HTML]{d2e7d6}{0.21} 
& \bf\colorbox[HTML]{d2e7d6}{0.14} 
& 0.53 
& \colorbox[HTML]{c0e7f6}{\underline{0.58}} 
& 0.45 
& \bf\colorbox[HTML]{d2e7d6}{0.43} \\
~~~-\textit{Llama} 
& \colorbox[HTML]{c0e7f6}{\underline{0.20}} 
& \bf\colorbox[HTML]{d2e7d6}{0.22} 
& \colorbox[HTML]{c0e7f6}{\underline{0.21}} 
& \colorbox[HTML]{c0e7f6}{\underline{0.13}} 
& \colorbox[HTML]{c0e7f6}{\underline{0.54}} 
& \bf\colorbox[HTML]{d2e7d6}{0.62} 
& \colorbox[HTML]{c0e7f6}{\underline{0.49}} 
& 0.35 \\
~~~-\textit{Gemma} 
& 0.19 
& 0.13 
& 0.16 
& 0.11 
& 0.40 
& 0.38 
& 0.40 
& 0.24 \\
~~~-\textit{Qwen} 
& \colorbox[HTML]{c0e7f6}{\underline{0.20}} 
& \colorbox[HTML]{c0e7f6}{\underline{0.20}} 
& \bf\colorbox[HTML]{d2e7d6}{0.21} 
& \colorbox[HTML]{c0e7f6}{\underline{0.13}} 
& \bf\colorbox[HTML]{d2e7d6}{0.59} 
& 0.57 
& \bf\colorbox[HTML]{d2e7d6}{0.52} 
& \bf\colorbox[HTML]{d2e7d6}{0.43} \\
~~~-\textit{Falcon} 
& \colorbox[HTML]{c0e7f6}{\underline{0.20}} 
& \colorbox[HTML]{c0e7f6}{\underline{0.20}} 
& \bf\colorbox[HTML]{d2e7d6}{0.21} 
& \bf\colorbox[HTML]{d2e7d6}{0.14} 
& \colorbox[HTML]{c0e7f6}{\underline{0.54}} 
& \colorbox[HTML]{c0e7f6}{\underline{0.57}} 
& 0.48
& \colorbox[HTML]{c0e7f6}{\underline{0.37}} \\
\hline
\end{tabular}}
\caption{Performance of Prompting methods across IBE-1 to IBE-4 settings in terms of macro F1 (MF) score. \colorbox[HTML]{d2e7d6}{\textbf{Green}} and \colorbox[HTML]{c0e7f6}{\underline{Blue}} indicate best and second-best performance, respectively.}
\label{table:prompting_ibe}
\end{table}

\begin{table}[t]
\centering
\footnotesize
\setlength{\tabcolsep}{1pt}
\renewcommand{\arraystretch}{1}
\resizebox{0.8\columnwidth}{!}{%
\begin{tabular*}{\columnwidth}{@{\extracolsep{\fill}}lcccccc@{}}
\hline
Dataset ($\xrightarrow{}$) & \multicolumn{3}{c}{\textbf{LIAR-RAW}} & \multicolumn{3}{c}{\textbf{RAW-FC}} \\
\cmidrule(lr){2-4} \cmidrule(lr){5-7}
Method ($\downarrow$) & MP & MR & MF1 & MP & MR & MF1 \\
\hline
\textsc{TBE-3} & 0.55 & 0.54 & 0.54 & 0.88 & 0.88 & 0.88 \\
~~~-\textit{w/o Supp. just.} & \colorbox[HTML]{c0e7f6}{\underline{0.38}} & \colorbox[HTML]{c0e7f6}{\underline{0.38}} 
& \colorbox[HTML]{c0e7f6}{\underline{0.37}}  
& \colorbox[HTML]{c0e7f6}{\underline{0.77}} 
& \colorbox[HTML]{c0e7f6}{\underline{0.78}} 
& \colorbox[HTML]{c0e7f6}{\underline{0.77}} \\

~~~-\textit{w/o Ref. just.} & \bf\colorbox[HTML]{d2e7d6}{0.49} & \bf\colorbox[HTML]{d2e7d6}{0.50} 
& \bf\colorbox[HTML]{d2e7d6}{0.49} 
& \bf\colorbox[HTML]{d2e7d6}{0.80} 
& \bf\colorbox[HTML]{d2e7d6}{0.80} 
& \bf\colorbox[HTML]{d2e7d6}{0.80} \\

~~~-\textit{w/o Both just.} & 0.26 & 0.26 & 0.24 & 0.46 & 0.46 & 0.46 \\

\hline
\end{tabular*}}
\caption{Ablation study showing classification performance on the LIAR-RAW (\textit{XLNet-L\textsubscript{Llama}}) and RAW-FC (\textit{RoBERTa-L\textsubscript{Llama}}) datasets. ``w/o Supp. just.'' indicates that only refuting justifications were passed to the model; ``w/o Ref. just.'' passes only supporting justifications; and ``w/o Both just.'' uses the claim alone without any justification. Here, we have reported the average values of three random seeds. \colorbox[HTML]{d2e7d6}{\textbf{Green}} and \colorbox[HTML]{c0e7f6}{\underline{Blue}} indicate best and second-best performance, respectively.}
\label{tab:ablation_study}
\end{table}

\subsection{Observations from ablation study:}
\label{sec:Detailed_observations_from_ablation_study}
In this section, we reported our observations from the ablation study. We used the best-performing models (\textit{XLNet-L\textsubscript{Llama}} for LIAR-RAW and \textit{RoBERTa-L\textsubscript{Llama}} for RAW-FC) on both datasets and removed the individual justification (supporting and refuting) components during training to gauge their impact. The results we obtained for different training scenarios are reported in Table \ref{tab:ablation_study}. Apart from that, we also segmented the test set into six parts based on the number of evidence pieces each claim had and calculated their performances. The segment-wise performance scores are reported in Table \ref{tab:liarraw_cluster_classification} and Table \ref{tab:rawfc_cluster_classification}, respectively. We observed the following:

\begin{itemize}

    \item When supporting justifications were removed, macro-F1 ($MF1$) scores dropped sharply by 31.48\% for LIAR-RAW and 12.50\% for RAW-FC. In contrast, the removal of refuting justifications had a milder but still noticeable impact, where $MF1$ decreased by 9.26\% and 9.09\% for LIAR-RAW and RAW-FC, respectively. When both supporting and refuting justifications were excluded, model performance degraded drastically, with $MF1$ decreasing by 55.56\% for LIAR-RAW and 47.73\% for RAW-FC. These findings emphasised that removing the supporting justifications had a more adversarial impact than removing refuting justifications, and removing both had the highest adversarial impact. 

    \item While investigating the behaviour with a varying number of evidence pieces, we observed the following. On LIAR-RAW, performance improved when a moderate amount of evidence was available, peaking at 6--20 evidence pieces ($MF1$: 0.61). Beyond this range, adding more evidence hurt performance, and $MF1$ dropped by \(29.50\%\downarrow\) for `$>50$' evidence pieces ($MF1$: 0.43), indicating sensitivity to evidence overload and likely exposure to more contradictory or less relevant statements. In contrast, RAW-FC benefited from richer evidence and achieved its best performance at 11--20 evidence pieces ($MF1$: 0.94). Even with `$>50$' evidence pieces, performance declined only slightly to $MF1$: 0.87 (\(\sim7.45\%\downarrow\)), suggesting that the predictor remained comparatively robust to larger evidence sets on RAW-FC.

\end{itemize}

\begin{table}[t]
\centering
\scriptsize
\setlength{\tabcolsep}{4pt}

\begin{subtable}[t]{0.46\columnwidth}
\centering
\begin{tabular}{p{1cm}ccc}
\hline
\textbf{No. Evi.} & \textbf{MP} & \textbf{MR} & \textbf{MF1} \\
\hline
0      & 0.47 & 0.51 & 0.48 \\
1      & 0.53 & 0.49 & 0.49 \\
2--5   & \colorbox[HTML]{c0e7f6}{\underline{0.58}} & \colorbox[HTML]{c0e7f6}{\underline{0.59}} & \colorbox[HTML]{c0e7f6}{\underline{0.58}} \\
6--20  & \colorbox[HTML]{d2e7d6}{\textbf{0.62}} & \colorbox[HTML]{d2e7d6}{\textbf{0.60}} & \colorbox[HTML]{d2e7d6}{\textbf{0.61}} \\
21--50 & 0.54 & 0.50 & 0.48 \\
$>$50  & 0.45 & 0.48 & 0.43 \\
\hline
\end{tabular}
\subcaption{LIAR-RAW}
\label{tab:liarraw_cluster_classification}
\end{subtable}
\hspace{0.05\columnwidth} 
\begin{subtable}[t]{0.46\columnwidth}
\centering
\begin{tabular}{p{1cm}ccc}
\hline
\textbf{No. Evi.} & \textbf{MP} & \textbf{MR} & \textbf{MF1} \\
\hline
4--5   & 0.83 & 0.91 & 0.84 \\
6--10  & 0.87 & 0.85 & 0.86 \\
11--20 & \colorbox[HTML]{d2e7d6}{\textbf{0.93}} & \colorbox[HTML]{d2e7d6}{\textbf{0.95}} & \colorbox[HTML]{d2e7d6}{\textbf{0.94}} \\
21--50 & 0.85 & 0.86 & 0.85 \\
$>$50  & \colorbox[HTML]{c0e7f6}{\underline{0.89}} & \colorbox[HTML]{c0e7f6}{\underline{0.88}} & \colorbox[HTML]{c0e7f6}{\underline{0.87}} \\
\hline
\end{tabular}
\subcaption{RAW-FC}
\label{tab:rawfc_cluster_classification}
\end{subtable}

\caption{Veracity detection performance of best performing models with respect to the evidence count (\textbf{No. Evid.}).  For LIAR-RAW, XLNet fine-tuned on Llama-based entailed justifications and for RAW-FC, RoBERTa fine-tuned on Llama-based entailed justifications are the best performing models. Metrics $MP$, $MR$, $MF1$ represent precision, recall and macro-F1. \colorbox[HTML]{d2e7d6}{\textbf{Green}} and \colorbox[HTML]{c0e7f6}{\underline{Blue}} indicate best and second-best performance, respectively.}
\end{table}

\subsection{Observations from domain generalisation study:} 
\label{sec:domain_gen}

As mentioned earlier, we evaluated the robustness of our learning paradigm, i.e., \textbf{TBE-3}, for multimodal and multilingual configurations. In the multimodal settings, we considered QwenVL \cite{bai2025qwen2}, Paligemma \cite{steiner2024paligemma}, and Idefics3 \cite{laurenccon2024building} (see Table \ref{tab:LLM_VLLM_Details} for details in the appendix) as GLMs. Similarly, in the multilingual settings, we considered the ELMs XLM-R \cite{conneau-etal-2020-unsupervised} and mBERT \cite{devlin-etal-2019-bert}. All models were trained using quantized low-rank adapters (QLoRA \cite{dettmers2023qlora}) for parameter-efficient fine-tuning. Further, we used the $4$-bit quantized versions of the GLMs, as they are memory-efficient yet provide comparable performance to their full-precision counterparts. We reported the macro-F1 scores ($MF1$) for the multimodal and multilingual datasets in Table \ref{table:tbe2_factify_mocheg_verite} and Table \ref{tab:multilingual_mf1_refstyle}, respectively. We observed the following:



\begin{itemize}
    \item Across multimodal settings, ELMs showed clear gains, improving MF1 by up to 33.3\%$\uparrow$, 37.8\%$\uparrow$, and 14.9\%$\uparrow$ for Factify-2, MOCHEG, and VERITE, respectively. QLoRA-based GLM fine-tuning further improved performance, achieving the best results across datasets, with improvements of up to 36.8\%$\uparrow$, 54.1\%$\uparrow$, and 36.2\%$\uparrow$ for Factify-2, MOCHEG, and VERITE, respectively. These improvements indicated that entailment-based supervision enabled robust multimodal reasoning for automatic fact-checking.

    \item In multilingual settings, fine-tuning ELMs yielded limited improvements, especially on X-Fact, highlighting the difficulty of multilingual generalisation. This was mainly due to X-Fact’s much larger language coverage (25 languages) and its 7-class veracity label. In contrast, QLoRA-tuned GLMs consistently outperformed both baselines and ELMs, improving MF1 by 9.5\%$\uparrow$ and 19.7\%$\uparrow$ for X-Fact and RU22Fact, respectively. This suggested that, when efficiently adapted, generative models better captured multilingual semantics and evidence interactions than encoder-only counterparts.
\end{itemize}

\begin{table}[t]
\centering
\footnotesize
\setlength{\tabcolsep}{4pt}
\renewcommand{\arraystretch}{1.1}
\resizebox{0.9\columnwidth}{!}{%
\begin{tabular}{lccc}
\hline
Dataset ($\xrightarrow{}$) & \textbf{Factify-2} & \textbf{MOCHEG} & \textbf{VERITE} \\
\cmidrule(lr){2-2} \cmidrule(lr){3-3} \cmidrule(lr){4-4}
Method ($\downarrow$) & MF1 & MF1 & MF1 \\
\hline



\rule{0pt}{3ex}\textsc{Fine-tuning} &  &  &  \\
~~~-\textit{RoBERTa-L\textsubscript{QwenVL}} & \colorbox[HTML]{d2e7d6}{\textbf{0.76$^{***}$}} & 0.47 & \colorbox[HTML]{d2e7d6}{\textbf{0.54}$^{**}$} \\
~~~-\textit{RoBERTa-L\textsubscript{Paligemma}}   & 0.71$^{***}$ & 0.48 & 0.40 \\
~~~-\textit{RoBERTa-L\textsubscript{Idefics3}}   & \colorbox[HTML]{c0e7f6}{\underline{0.75$^{***}$}} & \colorbox[HTML]{d2e7d6}{\textbf{0.51}} & \colorbox[HTML]{c0e7f6}{\underline{0.50$^{*}$}} \\ 
~~~-\textit{XLNet-L\textsubscript{QwenVL}}   & \colorbox[HTML]{d2e7d6}{\textbf{0.76$^{***}$}} & 0.48 & 0.48\\
~~~-\textit{XLNet-L\textsubscript{Paligemma}}     & 0.72$^{***}$ & 0.48 & 0.40 \\
~~~-\textit{XLNet-L\textsubscript{Idefics3}}     & 0.74$^{***}$ & \colorbox[HTML]{c0e7f6}{\underline{0.50}} & 0.48 \\
\hline
\rule{0pt}{3ex}\textsc{QLoRA} & & & \\
~~~-\textit{Llama-8B\textsubscript{QwenVL}}    & \colorbox[HTML]{c0e7f6}{\underline{0.75$^{***}$}} & \colorbox[HTML]{d2e7d6}{\textbf{0.57}} & 0.64$^{***}$ \\
~~~-\textit{Llama-8B\textsubscript{Paligemma}}    & 0.73$^{***}$ & 0.56 & 0.50$^{*}$ \\
~~~-\textit{Llama-8B\textsubscript{Idefics3}}    & 0.75$^{***}$ & 0.55 & 0.59$^{***}$ \\
~~~-\textit{Mistral-7B\textsubscript{QwenVL}}    & \colorbox[HTML]{d2e7d6}{\textbf{0.78$^{***}$}} & \colorbox[HTML]{c0e7f6}{\underline{0.56}} & \colorbox[HTML]{d2e7d6}{\textbf{0.64$^{***}$}} \\
~~~-\textit{Mistral-7B\textsubscript{Paligemma}}    & 0.72$^{***}$ & 0.57 & 0.50$^{*}$ \\
~~~-\textit{Mistral-7B\textsubscript{Idefics3}}    & 0.75$^{***}$ & 0.56 & \colorbox[HTML]{c0e7f6}{\underline{0.61$^{***}$}} \\

\hline
\rule{0pt}{3ex}\textsc{Baselines} &  &  &  \\
-\textit{\cite{du2023team}}   & 0.57 & -- & -- \\
-\textit{\cite{cekinel-etal-2025-multimodal}}   & -- & (0.37) & -- \\
-\textit{\cite{papadopoulos2025red}}   & -- & -- & 0.47 \\

\hline
\end{tabular}}

\caption{Veracity prediction performance of claim for multi-modal fact-checking datasets. \colorbox[HTML]{d2e7d6}{\textbf{Green}} and \colorbox[HTML]{c0e7f6}{\underline{Blue}} indicate best and second-best performance, respectively. We report the average of Macro-F1 scores across three random seeds. We reproduced the baselines, original reported values are kept in parenthesis. `*' indicates statistically significant improvements, $^{*}: p<0.05$, $^{**}: p<0.01$, and $^{***}: p<0.001$.
}
\label{table:tbe2_factify_mocheg_verite}
\end{table}

\begin{table}[t]
\centering
\footnotesize
\setlength{\tabcolsep}{5pt}
\renewcommand{\arraystretch}{1.1}
\resizebox{0.8\columnwidth}{!}{%
\begin{tabular}{lcc}
\hline
Dataset ($\xrightarrow{}$) & \textbf{X-Fact} & \textbf{Ru22fact} \\
\cmidrule(lr){2-2} \cmidrule(lr){3-3}
Method ($\downarrow$) & MF1 & MF1 \\
\hline

\rule{0pt}{3ex}\textsc{Fine-Tuning} & &  \\
~~~-\textit{XLMR\textsubscript{Gemma}}     & 0.40 & 0.65 \\
~~~-\textit{XLMR\textsubscript{Llama}} & 0.42$^{*}$ & 0.65 \\
~~~-\textit{XLMR\textsubscript{Mistral}}       & \colorbox[HTML]{d2e7d6}{\textbf{0.44}$^{***}$} & 0.67\\
~~~-\textit{XLMR\textsubscript{Qwen}}       & 0.41$^{***}$ & 0.68 \\
~~~-\textit{mBERT\textsubscript{Gemma}}     & 0.38 & 0.65 \\
~~~-\textit{mBERT\textsubscript{Llama}} & \colorbox[HTML]{c0e7f6}{\underline{0.42}} & \colorbox[HTML]{c0e7f6}{\underline{0.67}} \\
~~~-\textit{mBERT\textsubscript{Mistral}} & 0.38 & \colorbox[HTML]{d2e7d6}{\textbf{0.68}} \\
~~~-\textit{mBERT\textsubscript{Qwen}} & 0.37 & 0.66 \\

\hline

\rule{0pt}{3ex}\textsc{QLoRA} & &  \\
~~~-\textit{Gemma\textsubscript{Gemma}}     & 0.46$^{***}$ & \colorbox[HTML]{c0e7f6}{\underline{0.72}} \\
~~~-\textit{Llama\textsubscript{Llama}}     & \colorbox[HTML]{d2e7d6}{\textbf{0.46}$^{***}$} & 0.71 \\
~~~-\textit{Mistral\textsubscript{Mistral}} & \colorbox[HTML]{c0e7f6}{\underline{0.46}$^{***}$} & \colorbox[HTML]{d2e7d6}{\textbf{0.73}} \\
~~~-\textit{Qwen\textsubscript{Qwen}}       & 0.45$^{***}$ & 0.70 \\

\hline
\rule{0pt}{3ex}\textsc{Baselines} &  &  \\
-\textit{\citet{gupta2021x}}   & 0.42 & --  \\
-\textit{\citet{zeng2024ru22fact}}   & -- & 0.61  \\

\hline
\end{tabular}}

\caption{ Performance of veracity prediction for multilingual fact-checking datasets. \colorbox[HTML]{d2e7d6}{\textbf{Green}} and \colorbox[HTML]{c0e7f6}{\underline{Blue}} indicate best and second-best performance, respectively.  The reported Macro-F1 scores are averaged across three random seeds. We reproduced the baselines, original reported values are kept in parenthesis. `*' indicates statistically significant improvements, $^{*}: p<0.05$, $^{**}: p<0.01$, and $^{***}: p<0.001$.
}
\label{tab:multilingual_mf1_refstyle}
\vspace{-2mm}
\end{table}

\section{Conclusion:}

We drew the following conclusions from our experiments: 

\begin{itemize}
    \item Our study showed that incorporating entailment-based justifications improved fact-checking systems. Training language models with GLM-entailed justifications significantly surpassed the baseline macro-F1 scores, with improvements of up to \textbf{28.57\%} and \textbf{44.26\%} for LIAR-RAW and RAW-FC, respectively. Our approach of training with claim-evidence understanding (TBE-2) secured the second spot, with an increase of up to \textbf{16.39\%} on the RAW-FC dataset compared to the best baseline macro-F1. In contrast, the inference-based methods (IBEs) were unable to effectively utilise the justifications generated by GLMs and performed consistently poorly.

    \item In our subjective evaluation, we found that Llama-generated explanations were more informative, readable, and logically structured (details in Appendix \ref{sec:Detailed_observations_from_evaluation_of_explanations}). This aligned with the superior performance of models such as XLNet and RoBERTa (in TBE-3), which leveraged these explanations as inputs.

    \item The role of GLM-entailed justifications as a second step in TBE-3 was validated in the ablation study, where we observed that removing supporting or refuting justifications adversarially affected the scores.

    \item Linguistic analysis (see Appendix \ref{sec:Additional_observations_from_error_analysis}) also showed that ELMs focused more accurately on important factual words and signals that indicated whether a statement supported or contradicted a claim. However, for samples with borderline veracity labels, attention appeared scattered.

    \item Our domain generalisation study demonstrated that our learning paradigm (TBE-3) was robust across multimodal and multilingual settings. Overall, TBE-3 emerged as a scalable and resource-efficient paradigm for real-world fact-checking.
\end{itemize}

\bibliography{custom}

\appendix
\section*{Appendix}
\section{Additional dataset details}
\label{Additional_dataset_details}

The distribution of samples in the considered datasets is reported in Table \ref{dataset_statistics_detailed}. We chose \textbf{LIAR-RAW} and \textbf{RAW-FC} as they have been widely used in prior work, such as HiSS \cite{zhang2023towards}, FactLLaMa \cite{cheung2023factllama}, RAFTS \cite{yue2024retrieval}, L-Defence \cite{wang2024explainable}, and others \cite{wang2025end, xiong2025delphiagent}. While each sample in LIAR-RAW is annotated with one of six veracity labels, samples in RAW-FC follow a three-way labelling scheme. The datasets are open-sourced under the Apache 2.0 license. We used the training, validation, and test splits originally provided by \citet{yang-etal-2022-coarse}.

For multilingual evaluation, we considered X-FACT and RU22Fact and enriched the evidence by crawling claim-associated URLs. The full webpage content was extracted using Crawl4AI~\cite{crawl4ai2024}. Since the crawled documents were typically long and noisy, we segmented them into smaller textual chunks. To identify the most relevant evidence, we retrieved the top claim-matching chunk from each document using FAISS-based semantic search~\cite{douze2024faiss} with multilingual-e5 embeddings~\cite{wang2024multilingual}.

For multimodal evaluation, we considered Factify-2 \cite{DBLP:conf/defactify/SuryavardanMPCR23}, MOCHEG \cite{10.1145/3539618.3591879}, and VERITE \cite{papadopoulos2024verite}, which provide paired image–text inputs and are widely used benchmarks for multimodal misinformation detection. Factify-2 and MOCHEG differ in label granularity and reasoning complexity, while VERITE \cite{papadopoulos2024verite} is explicitly designed to reduce unimodal bias by enforcing joint visual–textual reasoning.

Since the test set of Factify-2 was not publicly available, we followed the approach of \cite{cekinel-etal-2025-multimodal} and sampled the same number of instances from the training set to create a validation set, and treated the official validation set as the test set for our experiments. For VERITE, we used the original three-way split and class definitions for our experiments.

We could not reproduce the RU22Fact baseline~\cite{zeng2024ru22fact} due to the unavailability of the official code, nor the MOCHEG baseline~\cite{cekinel-etal-2025-multimodal} because of its high computational cost (approximately 37 GPU hours per epoch). However, we successfully reproduced the Factify-2 baseline~\cite{du2023team} using a reduced maximum sequence length of 128 tokens to accommodate GPU memory constraints.


\begin{table}[t]
\centering
\footnotesize 
\setlength{\tabcolsep}{3.5pt} 
\renewcommand{\arraystretch}{1.1} 
\resizebox{\columnwidth}{!}{%
\begin{tabular}{|l|l|r|r|r|r|}
\hline
\textbf{Dataset} & \textbf{Classes} & \textbf{Count} & \textbf{Train} & \textbf{Val} & \textbf{Test} \\ 
\hline

\multirow{7}{*}{\makecell[l]{LIAR-RAW\\\cite{yang-etal-2022-coarse}}} 
 & True (T)          & 2,021 & 1,647 & 169 & 205 \\ 
 & Mostly-true (MT)  & 2,439 & 1,950 & 251 & 238 \\ 
 & Half-true (HT)    & 2,594 & 2,087 & 244 & 263 \\ 
 & Barely-true (BT)  & 2,057 & 1,611 & 236 & 210 \\ 
 & False (F)         & 2,466 & 1,958 & 259 & 249 \\ 
 & Pants-fire (PF)   & 1,013 & 812  & 115 & 86 \\ \hline
 & \textbf{Total}    & \textbf{12,590} &  &  &  \\ 
\hline

\multirow{4}{*}{\makecell[l]{RAW-FC\\\cite{yang-etal-2022-coarse}}} 
 & True (T)          & 695 & 561 & 67 & 67 \\ 
 & Half-true (HT)    & 671 & 537 & 67 & 67 \\ 
 & False (F)         & 646 & 514 & 66 & 66 \\ \hline
 & \textbf{Total}    & \textbf{2,012} & &  &  \\ 
\hline
\multirow{6}{*}{\makecell[l]{Factify-2\\\cite{DBLP:conf/defactify/SuryavardanMPCR23}}}
 & Support\_Multimodal       & 8,500 & 5,580 & 1,420 & 1,500 \\ 
 & Support\_Text             & 8,500 & 5,485 & 1,515 & 1,500 \\ 
 & Insufficient\_Multimodal  & 8,500 & 5,472 & 1,528 & 1,500 \\ 
 & Insufficient\_Text        & 8,500 & 5,494 & 1,506 & 1,500 \\ 
 & Refute                    & 8,500 & 5,469 & 1,531 & 1,500 \\ \hline
 & \textbf{Total}            & \textbf{42,500} & &  &  \\ 
\hline
\multirow{4}{*}{\makecell[l]{MOCHEG\\\cite{10.1145/3539618.3591879}}}
 & Supported         & 5,144 & 3,826 & 501 & 817 \\ 
 & Refuted           & 5,855 & 4,542 & 488 & 825 \\ 
 & Not Enough Info   & 4,602 & 3,301 & 501 & 800 \\ \hline
 & \textbf{Total}    & \textbf{15,601} & &  &  \\ 
\hline
\multirow{4}{*}{\makecell[l]{VERITE\\\cite{papadopoulos2024verite}}}
 & True              & 338 & 270 & 34 & 34 \\ 
 & MisCaptioned (MC) & 338 & 270 & 34 & 34 \\ 
 & Out-of-Context (OOC) & 324 & 271 & 33 & 34 \\ \hline
 & \textbf{Total}    & \textbf{1,000} &  &  &  \\ 
\hline

\multirow{8}{*}{\makecell[l]{X-Fact\\\cite{gupta-srikumar-2021-x}}}
 & False             & 12,536 & 7,515 & 977 & 4,044 \\ 
 & PT/M  & 7,311 & 4,359 & 553 & 2,399 \\ 
 & True              & 6,143 & 4,080 & 551  & 1,612 \\ 
 & Mostly True  & 1,947 & 1,380 & 194 & 373 \\ 
 & Mostly False & 1,536 & 848 & 114 & 574 \\ 
 & C/H  & 1,140 & 540 & 80 & 520 \\ 
 & Other             & 576 & 357 & 66 & 153 \\ \hline
 & \textbf{Total}    & \textbf{31,189} & &  &  \\ 
\hline
\multirow{4}{*}{\makecell[l]{RU22Fact\\\cite{zeng2024ru22fact}}}
 & Supported         & 10,081 & 6,836 & 1,079 & 2,166 \\ 
 & Refuted           & 4,651 & 3,299 & 450 & 902 \\ 
 & NEI (Not Enough Info) & 1,301 & 1,082 & 71 & 148 \\ \hline
 & \textbf{Total}    & \textbf{16,033} &  &  &  \\ 
\hline
\end{tabular}}
\caption{Distribution of samples across all datasets including train, validation, and test splits. \textbf{LIAR-RAW} and \textbf{RAW-FC} are monolingual English fact-checking benchmarks. \textbf{Factify-2}, \textbf{MOCHEG}, and \textbf{VERITE} are multimodal datasets with image-claim pairs. \textbf{X-Fact} is a multilingual dataset spanning 25 languages, where PT/M denotes Partly True/Mostly True and C/H denotes Completely/Highly False; the test split aggregates Indomain, Out-of-Domain (OOD), and Zeroshot evaluation sets. \textbf{RU22Fact} is a multilingual dataset covering Russia-Ukraine conflict claims in 4 languages.}
\label{dataset_statistics_detailed}
\end{table}

\section{Experimental set-up:}
\label{sec:exp_setup}

     \subsection{Evaluation metrics:} \label{subsubsec:eval_metrics} We used standard metrics such as macro-f1 ($MF1$) to evaluate our models. To assess explanation quality (based on concatenated supporting and refuting entailed justifications from TBE-3), we employed two types of strategies: (i) checking lexical overlap and semantic matching, and (ii) doing subjective evaluation by decoder-only GLMs. To check lexical overlap, we used several standard evaluation metrics such as ROUGE-1 ($R_1$), ROUGE-2 ($R_2$), ROUGE-L ($R_L$) \cite{lin-2004-rouge} and BLEU \cite{papineni-etal-2002-bleu}. While $R_1$ and $R_2$ measure the overlap of unigrams and bigrams between predicted and gold explanations, $R_L$ measures the longest common subsequence. BLEU, on the other hand, measures the precision of matching n-grams between predicted and gold explanations. To measure semantic matching between predicted and gold explanations, we used the BERT score \cite{zhang2019bertscore}. It generates contextual embeddings of predicted and gold explanations and calculates cosine similarity between them. To do the subjective evaluation, we prompted the considered decoder-only GLMs to assess the model explanations generated by each decoder-only GLM. The decoder-only GLMs were asked to assess across five dimensions (i) informativeness, (ii) logicality, (iii) objectivity, (iv) readability and (v) accuracy \cite{10.1145/3696410.3714532}. The prompt template used for the assessment is presented at ID 14 in Table \ref{tab:prompts}.

\subsection{Hyperparameter details:}We performed an extensive hyperparameter search that led to the optimal performance of our models. The list of hyperparameters for which we trained our models is presented in Table \ref{tab:hyperparams}. For the GLMs, we kept the temperature constant at `$0.001$' for consistency. We conducted all our experiments on NVIDIA A100 80GB GPU card. 

\section{Observations from model explainability:}
\label{sec:Detailed_observations_from_evaluation_of_explanations}
\begin{figure}[t]
    \begin{subfigure}[t]{0.48\columnwidth}
       \centering
        \includegraphics[width=\linewidth]{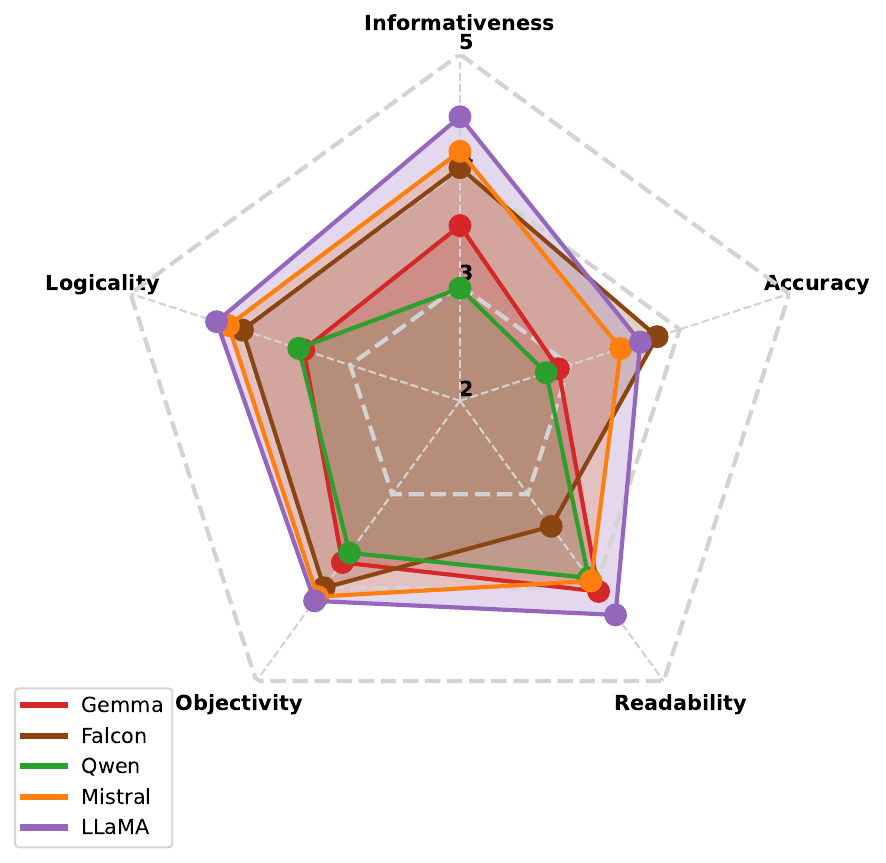}
        \caption{LIAR-RAW dataset}
        \label{fig:radar_lair} 
    \end{subfigure}
    \hfill
    \begin{subfigure}[t]{0.48\columnwidth}
        \centering
        \includegraphics[width=\linewidth]{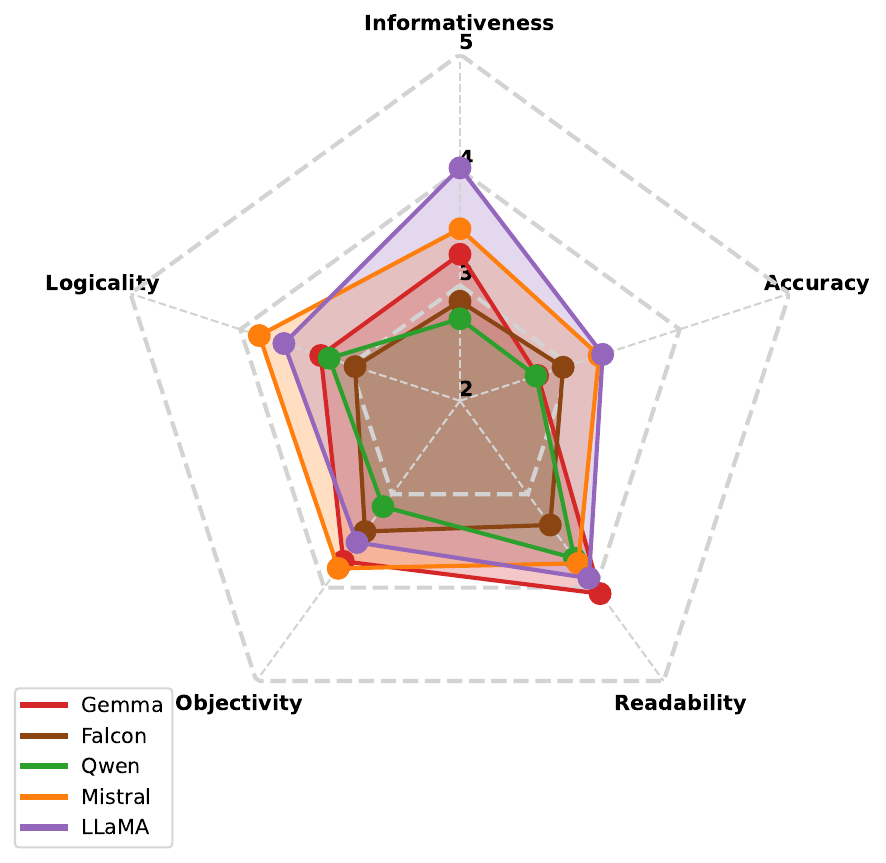}
        \caption{RAW-FC dataset}
        \label{fig:radar_rawfc}
    \end{subfigure}

    \caption{Average subjective evaluation scores across five models. Scores are reported for Informativeness (Info.), Accuracy (Acc.), Readability (Read.), Objectivity (Obj.), and Logicality (Logi.), as evaluated by generative language models. The raw values are reported in Table \ref{table:radar_scores} of Appendix.}
    \label{fig:radar_subjective_avg}
\end{figure}

In this section, we reported our observations from evaluating model explanations. The results of lexical-overlap and semantic-matching-based evaluations are reported in Table \ref{table:explanation}. Similarly, the average subjective evaluation ratings achieved by each GLM are presented in Figure \ref{fig:radar_subjective_avg}. Detailed scores are reported in Table \ref{table:radar_scores}. Some of the key findings are as follows:   

\begin{itemize}
    \item Falcon explanations achieved the highest $R_1$ score for both LIAR-RAW (\textbf{0.23}) and RAW-FC (\textbf{0.40}). This means they showed the maximum unigram overlap with the gold explanations. Similarly, Mistral and Falcon explanations achieved the highest $R_L$ score for LIAR-RAW (\textbf{0.14}). Mistral explanations also achieved the highest $R_L$ for RAW-FC (\textbf{0.18}). This indicates that these explanations exhibited the maximum overlap of the longest common subsequences with the gold explanations. Interestingly, we observed a small deviation in the $R_2$ scores. While explanations generated by Mistral, Llama, and Falcon achieved high $R_2$ scores for the LIAR-RAW dataset (\textbf{0.06}–\textbf{0.07}), Gemma achieved the highest score (\textbf{0.20}) for the RAW-FC dataset. This means their explanations showed the maximum bigram overlap with the gold explanations. 

    \item In evaluating GLMs, we observed consistent trends across both RAW-FC and LIAR-RAW datasets, with Llama and Mistral outperforming others in the majority of dimensions. Overall, Llama achieved the most balanced ratings across all dimensions, while consistently leading in informativeness. These observations correlated with the veracity prediction results, as RoBERTa and XLNet achieved the highest $MF1$ when trained with Llama-generated explanations. All models achieved higher absolute scores on LIAR-RAW compared to RAW-FC across dimensions. 
    
\end{itemize}

\begin{table}[t]
\centering 
\footnotesize 
\setlength{\tabcolsep}{1pt} 
\renewcommand{\arraystretch}{1}
\resizebox{0.8\columnwidth}{!}{%
\begin{tabular}{lcccccccccc}
\hline
& \multicolumn{5}{c}{\textbf{LIAR-RAW}} & \multicolumn{5}{c}{\textbf{RAW-FC}} \\ 
\cmidrule(lr){2-6} \cmidrule(lr){7-11} 
& $R_{1}$ & $R_{2}$ & $R_{L}$ & $BLEU$ & $BERT$ & $R_{1}$ & $R_{2}$ & $R_{L}$ & $BLEU$ & $BERT$ \\
\hline
\textit{Mistral} & 0.17 & \colorbox[HTML]{c0e7f6}{\underline{0.06}} & \colorbox[HTML]{d2e7d6}{\textbf{0.14}} & \colorbox[HTML]{d2e7d6}{\textbf{0.03}} & 0.03 & \colorbox[HTML]{c0e7f6}{\underline{0.39}} & \colorbox[HTML]{c0e7f6}{\underline{0.12}} & \colorbox[HTML]{d2e7d6}{\textbf{0.18}} & 0.04 & 0.02 \\
\textit{Llama}   & 0.19 & \colorbox[HTML]{d2e7d6}{\textbf{0.07}} & \colorbox[HTML]{c0e7f6}{\underline{0.12}} & \colorbox[HTML]{d2e7d6}{\textbf{0.03}} & 0.04 & \colorbox[HTML]{c0e7f6}{\underline{0.39}} & 0.11 & \colorbox[HTML]{c0e7f6}{\underline{0.17}} & 0.04 & 0.04 \\
\textit{Gemma}   & \colorbox[HTML]{c0e7f6}{\underline{0.20}} & 0.04 & 0.11 & \colorbox[HTML]{c0e7f6}{\underline{0.02}} & \colorbox[HTML]{d2e7d6}{\textbf{0.08}} & 0.19 & \colorbox[HTML]{d2e7d6}{\textbf{0.20}} & 0.11 & \colorbox[HTML]{d2e7d6}{\textbf{0.06}} & \colorbox[HTML]{d2e7d6}{\textbf{0.24}} \\
\textit{Qwen}    & 0.17 & \colorbox[HTML]{c0e7f6}{\underline{0.06}} & 0.10 & \colorbox[HTML]{c0e7f6}{\underline{0.02}} & 0.03 & 0.28 & 0.08 & 0.15 & 0.02 & \colorbox[HTML]{c0e7f6}{\underline{0.07}} \\
\textit{Falcon}  & \colorbox[HTML]{d2e7d6}{\textbf{0.23}} & \colorbox[HTML]{c0e7f6}{\underline{0.06}} & \colorbox[HTML]{d2e7d6}{\textbf{0.14}} & \colorbox[HTML]{d2e7d6}{\textbf{0.03}} & \colorbox[HTML]{c0e7f6}{\underline{0.05}} & \colorbox[HTML]{d2e7d6}{\textbf{0.40}} & \colorbox[HTML]{c0e7f6}{\underline{0.12}} & \colorbox[HTML]{c0e7f6}{\underline{0.17}} & \colorbox[HTML]{c0e7f6}{\underline{0.05}} & 0.02 \\
\hline
\end{tabular}}
\caption{Performance of explanation generation. \colorbox[HTML]{d2e7d6}{\textbf{Green}} and \colorbox[HTML]{c0e7f6}{\underline{Blue}} indicate best and second-best performance, respectively.}
\label{table:explanation}

\end{table}

\section{Linguistic insights:}
\label{sec:Additional_observations_from_error_analysis}

In this section, we analyzed the attention patterns of the best-performing models using justification-level visualizations (Tables \ref{tab:attention_liar_raw} and \ref{tab:attention_rawfc} in the appendix), highlighting the top \textbf{25}\% most-attended tokens.

\begin{itemize}
    \item For \textbf{true} claims, attention concentrated on affirmative and evidential phrases (e.g., “supported by multiple sources,” “credible evidence”) within supporting justifications, while refuting justifications received minimal focus. This indicated that the model aligned its predictions with semantically grounded support rather than spurious cues.

    \item For \textbf{half-true} claims, attention was distributed across conflicting linguistic signals (e.g., “accurate” vs. “inconsistencies undermine”), reflecting the model’s sensitivity to mixed evidence and ambiguity.

    \item For \textbf{false} claims, models strongly emphasized refuting terms (e.g., “false,” “disputes the claim,” “outdated story”), often ignoring weak or unsupported supporting justifications.
\end{itemize}

Overall, these patterns indicated that entailment-based language models effectively prioritized salient factual signals. However, borderline cases, such as half-true claims, remained challenging due to dispersed attention across conflicting narratives.

\section{Human Evaluation:}
\label{sec:Human Evalution}
\begin{table}[t]
\centering
\scriptsize
\setlength{\tabcolsep}{6pt}   
\renewcommand{\arraystretch}{1.1}
\begin{tabular}{lcccc}
\hline
\textbf{Criterion} & \multicolumn{2}{c}{\textbf{LIAR-RAW}} & \multicolumn{2}{c}{\textbf{RAW-FC}} \\
\cmidrule(lr){2-3} \cmidrule(lr){4-5}
& \textbf{FK} & \textbf{KA} & \textbf{FK} & \textbf{KA} \\
\hline
Informativeness & 0.25 & 0.48 & 0.21 & 0.20 \\
Accuracy        & 0.30 & 0.33 & 0.13 & 0.15 \\
Readability     & 0.46 & 0.54 & 0.07 & 0.07 \\
Objectivity     & 0.12 & 0.22 & 0.10 & 0.19 \\
Logicality      & 0.37 & 0.51 & 0.18 & 0.06 \\
\hline
\end{tabular}
\caption{Inter-annotator agreement (IAA) using Fleiss' Kappa (FK) and Krippendorff's Alpha (KA) for LIAR-RAW and RAW-FC. Low $\alpha$ (e.g., 0.073 for RAW-FC Readability) indicates agreement near chance level.}
\label{tab:iaa_results}
\vspace{-2mm}
\end{table}

We conducted a human evaluation of 40 randomly selected samples from LIAR-RAW and RAW-FC, with three independent annotators assessing explanations generated by the Llama model across five dimensions: \textit{Informativeness, Accuracy, Readability, Objectivity}, and \textit{Logicality}. Ratings were given on a Likert scale (1: Poor to 5: Excellent) consistent with prior GLMs evaluations, and inter-annotator agreement was measured using \textit{Fleiss’ Kappa} ($\kappa$) \cite{fleiss1971measuring}, \textit{Krippendorff’s Alpha} ($\alpha$) \cite{krippendorff2013content}.
For LIAR-RAW, agreement ranged from poor to moderate ($\kappa \approx 0.12$–$0.46$, $\alpha$ up to $0.54$), indicating reasonable consistency, particularly for readability and logicality.
In contrast, RAW-FC exhibited lower agreement overall ($\kappa \approx 0.076$–$0.21$, $\alpha$ from $0.06$ to $0.20$), reflecting greater subjectivity in judgments.
Detailed results are provided in Table~\ref{tab:iaa_results}.

\section{Limitations:}
\label{sec:Limitations}
Our work has some limitations as well. Firstly, our experiments are limited to open-source language models for reproducibility and resource efficiency. Secondly, we restricted ourselves to quantized models to expedite the domain generalization experiments. In future, one can check the generalizability with full precision and full-finetuning of GLMs (as well as utilize commercial GLMs). One can modify our approach to check its performance on a variety of larger fact-checking datasets like FEVER \cite{thorne-etal-2018-fever} series, fact-check-bench \cite{wang-etal-2024-factcheck}, etc. Lastly, we assumed a closed-domain fact-checking setup for reproducibility. Future work can consider open-domain fact-checking, i.e., retrieve evidence from an external source and test our hypothesis for generalization.

\begin{table}[h]
\centering
\footnotesize
\setlength{\tabcolsep}{4pt}
\renewcommand{\arraystretch}{1.05}
\resizebox{0.7\columnwidth}{!}{%
\begin{tabular}{l l l}
\hline
\textbf{Models} & \textbf{Parameters} & \textbf{Values} \\
\hline
\multirow{4}{*}{\textbf{ELMs}} 
& Learning rate & \{2e-6, 2e-5, 1e-5\} \\
& Optimizer & AdamW, Adam \\
& Batch size & 8, 16 \\
& Patience (Early stop) & 2, 3 \\
\midrule
\multirow{4}{*}{\textbf{GLMs}} 
& lora\_rank & 8 \\
& epoch & 3 \\
& Learning rate & \{1e-5, 1e-4\} \\
& lr\_scheduler\_type & cosine \\
& bf16 & true \\

\hline
\end{tabular}}
\caption{Hyperparameters explored during model training and evaluation.}
\label{tab:hyperparams}
\end{table}

\begin{table}[h]
\centering
\footnotesize
\setlength{\tabcolsep}{1pt}
\renewcommand{\arraystretch}{1.1}
\resizebox{\columnwidth}{!}{%
\begin{tabular}{lcr}
\hline
\textbf{Model} & \textbf{Max. length} & \textbf{Version} \\ \hline
\textsc{ELMs} \\
~~~-\textit{RoBERTa-L} & 1,024 & \texttt{FacebookAI/roberta-large}\\
~~~-\textit{XLNet-L} & 1,024 & \texttt{xlnet/xlnet-large-cased}\\ \hdashline
\textsc{Multilingual ELMs} \\
~~~-\textit{mBERT} & 512 & \texttt{google-bert/bert-base-multilingual-uncased}\\
~~~-\textit{XLM-R} & 1,024 & \texttt{FacebookAI/xlm-roberta-large}\\
~~~-\textit{multilingual-e5} & 512 & \texttt{intfloat/multilingual-e5-large-instruct}\\ \hdashline
\textsc{GLMs} \\
~~~-\textit{Mistral} & 32k  & \texttt{mistralai/Mistral-7B-Instruct-v0.3} \\
~~~-\textit{Llama} & 128k & \texttt{meta-llama/Llama-3.1-8B-Instruct} \\ 
~~~-\textit{Gemma} & 8k & \texttt{google/gemma-7b-it} \\
~~~-\textit{Qwen} & 1M & \texttt{Qwen/Qwen2.5-7B-Instruct-1M} \\ 
~~~-\textit{Falcon} & 8k & \texttt{tiiuae/Falcon3-7B-Instruct} \\ \hdashline
\textsc{Vision GLMs} \\
~~~-\textit{QwenVL} & 10k & \texttt{Qwen/Qwen2.5-VL-7B-Instruct}\\
~~~-\textit{Paligemma} & 8k & \texttt{google/paligemma2-3b-pt-448}\\
~~~-\textit{Idefics3} & 4k & \texttt{leon-se/Idefics3-8B-Llama3-bnb\_nf4}\\

\hline

\end{tabular}}

\caption{Language model details. Here, `Max. length' denotes the maximum supported sequence length allowed by the particular model. Rest of the parameters are directly taken from default settings of Llama Factory (for GLMs)\cite{zheng2024llamafactory}}
\label{tab:LLM_VLLM_Details}
\vspace{-2mm}
\end{table}

\onecolumn
\begin{center}
\small
\begin{longtable}{p{0.05\linewidth}p{0.20\linewidth}p{0.70\linewidth}}

\toprule
\textbf{ID} & \textbf{Caption} & \textbf{Prompt} \\
\midrule
\endfirsthead

\multicolumn{3}{c}{\tablename\ \thetable\ -- Continued from previous page} \\
\toprule
\textbf{ID} & \textbf{Caption} & \textbf{Prompt} \\
\midrule
\endhead

\midrule
\multicolumn{3}{r}{\small Continued on next page} \\
\bottomrule
\endfoot

\bottomrule
\endlastfoot

1 & IBE-1 (LIAR-RAW): Prompt used to predict the veracity of the claim using zero-shot prompting based on the given evidence sentences and the claim. &
\textbf{System Prompt:} "You are a fact-checking assistant. You are given a claim along with evidence sentences." "Your task is to determine the veracity of the claim based on the provided evidence sentences.\textbackslash n \textbackslash n". "Output your answer in exactly the following format:\textbackslash n" "Claim Veracity: [label]\textbackslash n \textbackslash n" "Answer with one word: true, false, half-true, mostly-true, barely-true, or pants-fire.\textbackslash n \textbackslash n" "Label Definitions:\textbackslash n" " true: The claim is completely supported by the evidence and is verified as accurate.\textbackslash n". " false: The claim is clearly refuted by the evidence and is demonstrably incorrect.\textbackslash n". " half-true: The claim contains both accurate and inaccurate details, making it partially true.\textbackslash n". " mostly-true: The claim is mostly supported by the evidence, but contains some inaccuracies.\textbackslash n". " barely-true: The claim contains some truth but is mostly misleading or inaccurate.\textbackslash n". " pants-fire: The claim is completely false and outrageously ridiculous."
\textbf{User Prompt:} f"Claim: {claim}\textbackslash n". f"Evidence: {evidence}\textbackslash n \textbackslash n". "Based on the above evidence sentences, determine the veracity of the claim." "Answer using one word (true, false, half-true, mostly-true, barely-true, pants-fire) using the specified format."\\
\midrule
2 & IBE-1 (RAW-FC): Prompt used to predict the veracity of the claim using zero-shot prompting based on the given evidence sentences and the claim. &
\textbf{System Prompt:} "You are a fact-checking assistant. You are given a claim along with evidences sentences. ". "Your task is to determine the overall veracity of the claim based on the provided evidence sentences.\textbackslash n \textbackslash n". "Output your answer in exactly the following format:\textbackslash n". "Claim Veracity: [label]\textbackslash n \textbackslash n". "Answer with one word: true, false, or half.\textbackslash n \textbackslash n". "Label Definitions for RAWFC:\textbackslash n". "true: The claim is completely supported by the evidence and is verified as accurate.\textbackslash n". "false: The claim is clearly refuted by the evidence and is demonstrably in incorrect.\textbackslash n". "half: The claim contains both accurate and inaccurate details, making it partially true." \textbf{User Prompt:} "Claim: {claim}\textbackslash n". f"Evidence: {evidence}\textbackslash n \textbackslash n". "Based on the above evidence, determine the veracity of the claim. ". "Answer using one word (true, false, or half) using the specified format."\\
\midrule
3 & IBE-2 and TBE-2: First step, prompt used to generate an overall understanding of the claim based on the provided evidence sentences. &
\textbf{System Prompt:} You are a helpful assistant. Your job is to read a claim and the raw evidence provided for it, and then explain your overall understanding of the claim based on that evidence. Focus on the key points and how the evidence helps in understanding the claim. Be objective, neutral, and avoid repetition. \textbf{User Prompt:} f"Here is a claim: \textbackslash"{claim}\textbackslash" \textbackslash n \textbackslash n"
f"Here is the evidence: \textbackslash"{evidences}\textbackslash" \textbackslash n \textbackslash n". "Based on the provided claim and evidence, what is your overall understanding of the claim?" "Provide a brief overall understanding (not more than 150 words) that captures the key reasoning between them." \\
\midrule
4 & IBE-2 (LIAR-RAW): Second step, prompt using zero-shot prompting to directly predict the claim's veracity based on the overall understanding. &
\textbf{System Prompt:} "You are a fact-checking assistant. You are given a claim along with an 'Understanding' of the evidence sentences of the claim."
"Do not consider any other evidence. Your task is to determine the veracity of the claim based on the understanding.\textbackslash n\textbackslash n" "Your task is to determine the claim's veracity using one of the following six categories:\textbackslash n\textbackslash n". "1) pants-fire : The claim is ridiculously or outrageously untrue.\textbackslash n". "2) false : The claim is entirely incorrect and unsupported by evidence.\textbackslash n". "3) barely-true : The claim has a small element of truth but is mostly misleading.\textbackslash n". "4) half-true : The claim mixes true and false information in significant amounts.\textbackslash n". "5) mostly-true : The claim is generally correct but contains minor inaccuracies.\textbackslash n". "6) true : The claim is completely accurate and fully supported by evidence.\textbackslash n\textbackslash n"."Output your answer in exactly the following format:\textbackslash n". "Claim Veracity: [label]\textbackslash n\textbackslash n". "where [label] must be one of (pants-fire, false, barely-true, half-true, mostly-true, or true)." \textbf{User Prompt:} f"Claim: {claim}\textbackslash n" f"Understanding: {understanding}\textbackslash n\textbackslash n". "Based on the above understanding, determine the veracity of the claim. Answer using one word label from the following options:\textbackslash n". "pants-fire\textbackslash n" "false\textbackslash n" "barely-true\textbackslash n" "half-true\textbackslash n" "mostly-true\textbackslash n" "true\textbackslash n\textbackslash n" "Answer only with: Claim Veracity: [label]." \\
\midrule
5 & IBE-2 (RAW-FC): Second step, using zero-shot prompting to predict the claim's veracity from the overall understanding. &
\textbf{System Prompt:} "You are a fact-checking assistant. You are given a claim along with an 'Understanding' of the evidence sentences." "Do not consider any other evidence. Your task is to determine the overall veracity of the claim based on the understanding.\textbackslash n\textbackslash n" "Output your answer in exactly the following format:\textbackslash n" "Claim Veracity: [label]\textbackslash n\textbackslash n" "Answer with one word: true, false, or half." \textbf{User Prompt:} f"Claim: {claim}\textbackslash n" f"Understanding: {understanding}\textbackslash n\textbackslash n" "Label Definitions for RAWFC:\textbackslash n" "true: The claim is completely supported by the evidence and is verified as accurate.\textbackslash n" "false: The claim is clearly refuted by the evidence and is demonstrably incorrect.\textbackslash n" "half: The claim contains both accurate and inaccurate details, making it partially true.\textbackslash n\textbackslash n" "Based solely on the above understanding, determine the overall veracity of the claim." "Answer using one word (true, false, or half) using the specified format." \\

6 & IBE-3: First step, prompt used to generate an overall understanding of the claim based on the provided evidence sentences. &
\textbf{System Prompt:} You are a helpful assistant. Your job is to read a claim and the raw evidence provided for it, and then explain your overall understanding of the claim based on that evidence. Focus on the key points and how the evidence helps in understanding the claim. Be objective, neutral, and avoid repetition. \textbf{User Prompt:} f"Here is a claim: \textbackslash"{claim}\textbackslash"\textbackslash n\textbackslash n" f"Here is the evidence: \textbackslash"{evidences}\textbackslash"\textbackslash n\textbackslash n" "Based on the provided claim and evidence, what is your overall understanding of the claim?" "Provide a brief overall understanding (not more then 150 words) that captures the key reasoning between them." \\
\midrule
7 & IBE-3 (LIAR-RAW): Second step, prompt using Chain-of-Thought reasoning to predict the veracity of the claim from the overall understanding. &
\textbf{System Prompt:} "You are a fact-checking assistant. You are given a claim along with an 'Understanding' of the evidence sentences for the claim. " "Do not consider any other evidence. Your task is to determine the veracity of the claim based on the understanding.\textbackslash n\textbackslash n" "First, explain your reasoning step-by-step using only the understanding provided.\textbackslash n" "Then, determine the claim's veracity using one of the following six categories:\textbackslash n\textbackslash n" "1) pants-fire : The claim is ridiculously or outrageously untrue.\textbackslash n" "2) false : The claim is entirely incorrect and unsupported by evidence.\textbackslash n" "3) barely-true : The claim has a small element of truth but is mostly misleading.\textbackslash n" "4) half-true : The claim mixes true and false information in significant amounts.\textbackslash n" "5) mostly-true : The claim is generally correct but contains minor inaccuracies.\textbackslash n" "6) true : The claim is completely accurate and fully supported by evidence.\textbackslash n\textbackslash n" "Finally, output your answer in exactly the following format:\textbackslash n" "Claim Veracity: [label]\textbackslash n\textbackslash n" "where [label] must be one of (pants-fire, false, barely-true, half-true, mostly-true, or true)." \textbf{User Prompt:} f"Claim: {claim}\textbackslash n" f"Understanding: {understanding}\textbackslash n\textbackslash n" "First, explain your reasoning based only on the understanding provided above.\textbackslash n" "Then, determine the overall veracity of the claim using one of the following labels:\textbackslash n" "- pants-fire\textbackslash n" "- false\textbackslash n" "- barely-true\textbackslash n" "- half-true\textbackslash n" "- mostly-true\textbackslash n" "- true\textbackslash n\textbackslash n" "Finally, answer in the following format:\textbackslash n" "Claim Veracity: [label]" \\
\midrule
8 & IBE-3 (RAW-FC): Second step, prompt using Chain-of-Thought reasoning to predict the veracity of the claim from the overall understanding. &
\textbf{System Prompt:} "You are a fact-checking assistant. You are given a claim along with an 'Understanding' of the claim. " "Do not consider any other evidence. Your task is to determine the overall veracity of the claim based on the understanding.\textbackslash n\textbackslash n" "First, explain your reasoning in detail based only on the understanding.\textbackslash n" "Then output the claim veracity label in **exactly** the following format:\textbackslash n" "Claim Veracity: [label]\textbackslash n\textbackslash n" "Answer with one word: true, false, or half." \textbf{User Prompt:} f"Claim: {claim}\textbackslash n" f"Understanding:{understanding}\textbackslash n\textbackslash n" "Label Definitions:\textbackslash n" "true: The claim is completely supported by the evidence and accurate.\textbackslash n" "false: The claim is clearly refuted by the evidence and incorrect.\textbackslash n" "half: The claim contains both accurate and inaccurate details, making it partially true.\textbackslash n\textbackslash n" "First, explain your reasoning step-by-step using only the understanding above.\textbackslash n" "Then, based on your reasoning, provide the final claim veracity in the format:\textbackslash n" "Claim Veracity: [true/false/half] \\
\midrule
9 & IBE-4 and TBE-3: First step prompt used to classify each evidence sentence as supporting or refuting the claim. &
\textbf{System Prompt:} You are an evidence classifier that determines whether a sentence supports or refutes a given claim. \textbf{User Prompt:} Task: Analyze the sentence below and classify its relationship to the given claim. Classification Rules: 1. SUPPORT - The sentence provides evidence that supports the claim. 2. REFUTE- The sentence provides evidence that contradicts or refutes the claim. Claim: {claim} Label for Claim: {label} Sentence to classify: {sent} Carefully analyze the sentence and respond with: 1. A classification (SUPPORT or REFUTE). \\
\midrule
10 & IBE-4 and TBE-3, Second step prompt used to generate a supporting justification based on the supporting evidence sentences. &
\textbf{System Prompt:} You have been specially designed to perform abductive reasoning for the fake news detection task and justification generation. \textbf{User Prompt:} Given a claim: {claim}\textbackslash n a veracity label True, analyze the statement and provide a concise, well-reasoned explanation (150 words or fewer) based on the provided evidence. Do not mention the label '{label}' in your response. Focus only on key supporting factors and logical reasoning. Evidence: {entry['supporting\_evidence']} \\
\midrule
11 & IBE-4 and TBE-3, Second step prompt used to generate a refuting justification based on the refuting evidence sentences. &
\textbf{System Prompt:} You have been specially designed to perform abductive reasoning for the fake news detection task and justification generation. \textbf{User Prompt:} Given a claim: {claim}\textbackslash n a veracity label False, analyze the statement and provide a concise, well-reasoned explanation (150 words or fewer) based on the provided evidence. Do not mention the label '{label}' in your response. Focus on key factors that challenge the claim while maintaining clarity and coherence. Evidence: {entry['refuting\_evidence']} \\

12 & IBE-4 (LIAR-RAW): Third step prompt used to predict the veracity of the claim using the supporting and refuting justifications. &
\textbf{System Prompt:} "You are a fact-checking assistant. You are given a claim along with justifications: a 'Support Justification' and a 'Refute Justification'." "Do not consider any other evidence. Your task is to compare these justifications and determine the veracity of the claim based on justification.\textbackslash n\textbackslash n" "Your task is to determine the claim's veracity using one of the following six categories:\textbackslash n\textbackslash n" "1) pants-fire : The claim is ridiculously or outrageously untrue.\textbackslash n" "2) false : The claim is entirely incorrect and unsupported by evidence.\textbackslash n" "3) barely-true : The claim has a small element of truth but is mostly misleading.\textbackslash n" "4) half-true : The claim mixes true and false information in significant amounts.\textbackslash n" "5) mostly-true : The claim is generally correct but contains minor inaccuracies.\textbackslash n" "6) true : The claim is completely accurate and fully supported by evidence.\textbackslash n\textbackslash n" "Output your answer in exactly the following format:\textbackslash n" "Claim Veracity: [label]\textbackslash n\textbackslash n" "where [label] must be one of (pants-fire, false, barely-true, half-true, mostly-true, or true)." \textbf{User Prompt:} f"Claim: {claim}\textbackslash n" f"Support Justification 1: {true\_just}\textbackslash n" f"refute Justification 2: {false\_just}\textbackslash n\textbackslash n" "Based on the justifications above, determine the best label from the following options:\textbackslash n" "pants-fire\textbackslash n" " false\textbackslash n" " barely-true\textbackslash n" " half-true\textbackslash n" " mostly-true\textbackslash n" " true\textbackslash n\textbackslash n" "Answer only with: Claim Veracity: [label]." \\
\midrule

13 & IBE-4 (RAW-FC): Third step prompt used to predict the veracity of the claim using the supporting and refuting justifications. &
\textbf{System Prompt:} "You are a fact-checking assistant. You are given a claim along with two generated justifications: a 'Support Justification' and a 'Refute Justification'. " "Do not consider any other evidence. Your task is to compare these justifications and determine the overall veracity of the claim based on justification.\textbackslash n\textbackslash n" "Output your answer in exactly the following format:\textbackslash n" "Claim Veracity: [label]\textbackslash n\textbackslash n" "Answer with one word: true, false, or half." \textbf{User Prompt:} f"Claim: {claim}\textbackslash n" f"Support Justification: {support\_just}\textbackslash n" f"Refute Justification: {refute\_just}\textbackslash n\textbackslash n" "Label Definitions for RAWFC:\textbackslash n" "true: The claim is completely supported by the evidence and is verified as accurate.\textbackslash n" "false: The claim is clearly refuted by the evidence and is demonstrably incorrect.\textbackslash n" "half: The claim contains both accurate and inaccurate details, making it partially true.\textbackslash n\textbackslash n" "Based on the above justifications, determine the overall veracity of the claim. " "Answer using one word (true, false, or half) using the specified format." \\
\midrule
14 & Prompt sample for subjective evaluation of generated explanation. &
\textbf{System Prompt:} "" Your task is to evaluate a model-generated explanation ("generated") against a reference explanation ("gold") across five specific criteria. For each criterion, assign an integer score from 1 to 5 (inclusive): 1 = very poor, 2 = poor, 3 = average, 4 = good, 5 = excellent"". \textbf{User Prompt:} """**Instructions:** - Only return a **strict JSON object** in the exact format shown below. - Do NOT include any explanation, comments, or additional text. - Do NOT use floating point numbers or non-integer values. - If you are unsure, always pick the conservative score. **Criteria:** 1. Informativeness – Does it add meaningful context or background? 2. Accuracy – Does it reflect the correct meaning based on the reference? 3. Readability – Is it grammatically correct and easy to understand? 4. Objectivity – Is it neutral and free of emotional language? 5. Logicality – Does the reasoning follow a coherent and sound process? **Return JSON (strict format):** {{"Informativeness": int, "Accuracy": int, "Readability": int, "Objectivity": int, "Logicality": int}} Expected explanation: \textbackslash"\textbackslash"\textbackslash"{expected}\textbackslash"\textbackslash"\textbackslash" Actual explanation:
\textbackslash"\textbackslash"\textbackslash"{actual}\textbackslash"\textbackslash"\textbackslash" """ \\
\midrule
\caption{Collection of prompt templates used in TBEs and IBEs experiments, covering various stages such as claim understanding, evidence classification, justification generation, and veracity prediction. The final prompt is used for subjective evaluation of generated explanations.}
\label{tab:prompts}\\

\end{longtable}
\end{center}

\begin{center}
\small
\begin{longtable}{p{0.01\linewidth}p{0.93\linewidth}}
\hline
\textbf{ID}  & \textbf{Content with Attention Scores} \\
\hline
\endfirsthead

\hline
\textbf{ID}  & \textbf{Content with Attention Scores} \\
\hline
\endhead

\hline
\multicolumn{2}{r}{\small Continued on next page} \\
\hline
\endfoot

\hline
\endlastfoot
1 &
\textbf{Claim:} `` \textcolor{medred}{Police} \textcolor{medred}{pepper-sprayed} a double amputee and removed his prosthetic legs during a June \textcolor{medblue}{21,} \textcolor{medblue}{2020,} protest against police brutality in \textcolor{medred}{Columbus,} \textcolor{medred}{Ohio.}'' \quad \textbf{Gold Label:} ``half'' \quad \textbf{Predicted Label:} ``half'' \newline
\textbf{Support Justifications:} ``Based \textcolor{medred}{on} the provided \textcolor{medblue}{evidence,} \textcolor{medred}{I} \textcolor{medred}{conclude} \textcolor{medred}{that} \textcolor{medblue}{the} \textcolor{medred}{statement} \textcolor{blue}{is} TRUE. \textcolor{medblue}{The} evidence consists \textcolor{medred}{of} multiple reports, \textcolor{medblue}{videos,} and witness accounts from various sources, including Twitter users, news outlets, and the Columbus mayor. These accounts consistently describe the incident where a double amputee was \textcolor{medred}{pepper-sprayed} by police and had his prosthetic legs removed during a protest in Columbus, \textcolor{medred}{Ohio,} on June \textcolor{medblue}{21,} 2020. The details of \textcolor{medblue}{the} \textcolor{medred}{incident,} including \textcolor{medblue}{the} removal of the prosthetic legs and the officer's \textcolor{medred}{identity,} are corroborated across multiple sources, lending \textcolor{medred}{credibility} \textcolor{blue}{to} \textcolor{medblue}{the} reports. The widespread outrage and public attention to \textcolor{medblue}{the} incident further \textcolor{medred}{support} \textcolor{medblue}{the} \textcolor{medred}{validity} \textcolor{medred}{of} \textcolor{medblue}{the} \textcolor{medred}{statement.}'' \newline
\textbf{Refute Justifications:} `` \textcolor{medblue}{The} \textcolor{medred}{claim} \textcolor{medred}{that} police \textcolor{medred}{pepper-sprayed} a double amputee and removed his prosthetic legs during a June \textcolor{medblue}{21,} \textcolor{medblue}{2020,} protest against police brutality in \textcolor{medred}{Columbus,} \textcolor{medred}{Ohio,} \textcolor{blue}{is} false. \textcolor{medblue}{This} \textcolor{medred}{conclusion} \textcolor{blue}{is} supported \textcolor{medred}{by} \textcolor{medblue}{multiple} lines of \textcolor{medred}{evidence.} \textcolor{medred}{Firstly,} the Columbus Police Department disputes the \textcolor{medred}{claim,} stating that the individual attacked the officers and was carried away by \textcolor{medred}{protesters.} \textcolor{medblue}{This} \textcolor{blue}{is} corroborated by police body camera footage and a pole-mounted video camera at \textcolor{medblue}{the} \textcolor{medred}{scene,} which show the individual throwing a sign and a bottle of liquid at the \textcolor{medred}{officers.} \textcolor{medred}{Secondly,} there \textcolor{medred}{is} no footage of the prosthetic leg being removed \textcolor{medred}{from} the man, who returns a short time later with his leg \textcolor{medred}{attached,} according to police \textcolor{medred}{video.} \textcolor{medblue}{This} \textcolor{medred}{suggests} that the prosthetic leg was not taken by the \textcolor{medred}{police.} \textcolor{medred}{Thirdly,} witnesses from the scene describe the man crawling on his hands to get medical \textcolor{medred}{help,} and a group of protesters rushing the officer to get his leg \textcolor{medred}{back.} However, the police video shows that the protesters were the ones who pulled the man away from the officer, causing him to lose his prosthetic \textcolor{medred}{leg.} \textcolor{medred}{Lastly,} the police have evidence that shows the individual attacked the \textcolor{medred}{officers,} which contradicts the \textcolor{medred}{claim} that the police removed the prosthetic leg without \textcolor{medred}{provocation.} \textcolor{medblue}{In} \textcolor{medred}{conclusion,} the evidence \textcolor{medred}{suggests} that the \textcolor{medred}{claim} \textcolor{blue}{is} false, and the police did not remove the prosthetic leg \textcolor{medred}{from} the double amputee. \textcolor{medblue}{The} \textcolor{medred}{incident} was a result of a violent clash between the police and \textcolor{medred}{protesters,} and the man's prosthetic leg was lost due \textcolor{medred}{to} the actions of the \textcolor{medred}{protesters,} not the \textcolor{medred}{police.}'' \\
\midrule
2 &
\textbf{Claim:} ``\textcolor{medblue}{52} thoroughbred horses whose owner died from COVID-19 are destined for slaughter \textcolor{medred}{if} adopters are \textcolor{medred}{not} found for them \textcolor{medred}{soon.}'' \quad \textbf{Gold Label:} ``false'' \quad \textbf{Predicted Label:} ``false'' \newline
\textbf{Support Justifications:} ``\textcolor{medblue}{The} \textcolor{medred}{claim} \textcolor{blue}{is} likely true. \textcolor{medblue}{The} provided evidence \textcolor{medred}{suggests} that there \textcolor{medred}{is} an urgent need \textcolor{medred}{to} find new homes for 52 thoroughbred horses due \textcolor{medred}{to} the death of their \textcolor{medred}{owner.} \textcolor{medred}{This} \textcolor{medred}{implies} that the horses are indeed at risk of being sent to slaughter \textcolor{medred}{if} adopters are not found \textcolor{medred}{soon.} \textcolor{medblue}{The} fact that a circulating message is asking recipients \textcolor{blue}{to} help find new homes for the horses further supports the \textcolor{medred}{urgency} \textcolor{medred}{of} the \textcolor{medred}{situation.} \textcolor{medred}{Additionally,} the specific number of horses mentioned (52) and the breed (thoroughbred) provide context and \textcolor{medred}{credibility} \textcolor{blue}{to} the \textcolor{medred}{claim.} \textcolor{medblue}{Overall,} the evidence \textcolor{medred}{suggests} \textcolor{medred}{that} the \textcolor{medred}{claim} \textcolor{medred}{is} based on a genuine concern for the welfare of the horses and is not a fabricated or exaggerated \textcolor{medred}{story.}'' \newline
\textbf{Refute Justifications:} ``\textcolor{medblue}{The} \textcolor{medred}{claim} that \textcolor{medblue}{52} thoroughbred horses are destined for \textcolor{medred}{slaughter} \textcolor{medred}{if} adopters are not found for them \textcolor{medred}{soon} \textcolor{blue}{is} \textcolor{medred}{challenged} by several key \textcolor{medred}{factors.} \textcolor{medred}{Firstly,} the \textcolor{medred}{claim} has been circulating for several \textcolor{medred}{years,} with the original post dating back \textcolor{medred}{to} \textcolor{medblue}{2011.} According to credible reports on horse-related \textcolor{medred}{forums,} all 52 horses were actually \textcolor{medred}{rehomed,} with most going to family friends of the deceased \textcolor{medred}{owner.} \textcolor{medred}{This} \textcolor{medred}{suggests} that the claim \textcolor{medred}{is} not based on current \textcolor{medred}{events,} but rather a recycled and outdated \textcolor{medred}{story.} \textcolor{medred}{Furthermore,} the claim has undergone subtle changes over the \textcolor{medred}{years,} including the reason for the horses being in \textcolor{medred}{danger.} Initially, the post stated that the horses would be sent to a glue \textcolor{medred}{factory,} but more recently, the reason cited \textcolor{medred}{is} the owner's passing due \textcolor{medred}{to} \textcolor{medblue}{COVID-19.} \textcolor{medred}{However,} there \textcolor{medred}{is} no evidence \textcolor{medred}{to} suggest that the horses are currently in danger or that they are being considered for \textcolor{medred}{slaughter.} \textcolor{medred}{In} \textcolor{medred}{fact,} according \textcolor{medred}{to} TheHorse.com, all 52 horses were able \textcolor{medred}{to} find homes within a week of the initial post in \textcolor{medblue}{2011.} \textcolor{medblue}{This} contradicts the \textcolor{medred}{claim} that the horses are in imminent danger of being slaughtered. \textcolor{medred}{Therefore,} based on the available \textcolor{medred}{evidence,} \textcolor{medred}{it} \textcolor{medred}{appears} that the \textcolor{medred}{claim} \textcolor{blue}{is} not supported by \textcolor{medred}{facts} and \textcolor{medred}{is} likely a recycled and outdated \textcolor{medred}{story.}'' \\
\midrule
3 & 
\textbf{Claim:} ``\textcolor{medblue}{The} Irish Times `\textcolor{medred}{newspaper}' published \textcolor{medred}{a} column by Fintan `O' `Toole expressing `pity' for \textcolor{medblue}{the} United States over \textcolor{medblue}{U.S.} President Donald Trump' 's leadership during the COVID-19 pandemic '' \quad \textbf{Gold Label:} ``true'' \quad \textbf{Predicted Label:} ``true'' \newline
\textbf{Support Justifications:} `` \textcolor{medblue}{The} \textcolor{medred}{claim} \textcolor{medred}{suggests} \textcolor{medred}{that} Fintan O' `Toole wrote a column in The Irish Times expressing pity for the United States due to President Trump' 's handling of the pandemic. \textcolor{lightblue}{This} \textcolor{lightblue}{can} \textcolor{lightblue}{be} \textcolor{medblue}{verified} through \textcolor{medblue}{multiple} \textcolor{medblue}{sources.} \textcolor{medred}{Firstly,} Fintan `O' `Toole is a well-known columnist for The Irish Times, \textcolor{medred}{and} his opinions are widely \textcolor{medred}{respected.} Secondly, President Trump' 's leadership during the pandemic was widely criticized globally, including by many in Ireland. \textcolor{medred}{It} \textcolor{lightblue}{is} plausible that O' `Toole would express sympathy for the US in light of this \textcolor{medred}{criticism.} Furthermore, The Irish Times has a reputation for publishing high quality \textcolor{medred}{journalism,} and it is unlikely that they would publish a column without fact-checking its content. \textcolor{medred}{Therefore,} \textcolor{medred}{it is} reasonable \textcolor{lightblue}{to} conclude \textcolor{medred}{that} \textcolor{medblue}{the} \textcolor{medred}{claim} \textcolor{lightblue}{is} \textcolor{medred}{accurate.} \textcolor{medblue}{The} combination of O' `Toole' 's credibility as a columnist, the global criticism of Trump' 's leadership, and The Irish Times' reputation for quality journalism \textcolor{medred}{all} \textcolor{medred}{support} \textcolor{medblue}{the} \textcolor{medred}{validity} \textcolor{medred}{of} \textcolor{medblue}{the} \textcolor{medred}{claim}'' \newline
\textbf{Refute Justifications:} `` \textcolor{medblue}{The} \textcolor{medred}{claim} \textcolor{medred}{that} \textcolor{medblue}{`The} Irish Times `\textcolor{medred}{newspaper}' published a column by Fintan O' `Toole \textcolor{medred}{expressing} ``pity'' for \textcolor{medblue}{the} United States over \textcolor{medblue}{U.S.} President Donald Trump' 's leadership during the COVID-19 pandemic \textcolor{lightblue}{is} \textcolor{medred}{challenged} \textcolor{medred}{by} \textcolor{medblue}{the} content of the column \textcolor{medred}{itself.} While O' `Toole \textcolor{medred}{does express} \textcolor{medred}{sympathy} for \textcolor{medblue}{the} majority of Americans who did not vote for Trump, he does \textcolor{medred}{not} express \textcolor{medred}{a} sense of pity for \textcolor{medblue}{the} country as \textcolor{medred}{a} \textcolor{medred}{whole.} In fact, he suggests that the United States has historically evoked a wide range of emotions \textcolor{medred}{globally,} including love, hatred, \textcolor{medred}{fear,} and contempt. Furthermore, O' `Toole critiques Trump' 's \textcolor{medred}{leadership,} calling him an ``authoritarian and con man,'' and criticizes the president's handling of the pandemic, stating that he is ``actively promoting the spread of a fatal disease.'' The tone of the column is critical of Trump and his \textcolor{medred}{administration,} rather \textcolor{medred}{than} expressing pity for the \textcolor{medred}{country.} \textcolor{medred}{Therefore,} \textcolor{medblue}{the claim} \textcolor{medred}{appears} \textcolor{lightblue}{to} \textcolor{medred}{be} \textcolor{medred}{inaccurate.}'' \\
\hline
\\[0.1em]
\caption{Attention-Based Explanation Samples for Fact-Checking Predictions on RAWFC Datasets.}
\label{tab:attention_rawfc}
\end{longtable}
\end{center}

\begin{center}
\small
\begin{longtable}{p{0.01\linewidth}p{0.93\linewidth}}
\hline
\textbf{ID}  & \textbf{Content with Attention Scores} \\
\hline
\endfirsthead

\hline
\textbf{ID}  & \textbf{Content with Attention Scores} \\
\hline
\endhead

\hline
\multicolumn{2}{r}{\small Continued on next page} \\
\hline
\endfoot

\hline
\endlastfoot

1 & 
\textbf{Claim:} ``Foreign aid \textcolor{red}{is} less than 1 percent of \textcolor{blue}{our} federal \textcolor{red}{budget}.'' 
\quad \textbf{Gold Label:} ``true'' 
\quad \textbf{Predicted Label:} ``true'' 
\textbf{Support Justifications:} ``\textcolor{medblue}{The} \textcolor{blue}{statement} \textcolor{red}{that} foreign aid is less than 1 percent of \textcolor{red}{our} federal budget \textcolor{blue}{is} \textcolor{red}{supported by} \textcolor{medblue}{multiple} \textcolor{medblue}{sources.} According \textcolor{red}{to} various statements, foreign aid makes up less than one percent of \textcolor{medblue}{the} total federal budget, is \textcolor{red}{a} ``drop in \textcolor{medblue}{the} budgetary bucket,'' and amounts to far less than what most Americans think is spent on it. \textcolor{medblue}{Additionally,} \textcolor{red}{it} ``\textcolor{red}{s} stated that less than 1 percent of the 4 trillion federal budget goes to foreign aid, and Americans actually spend less than 1 percent of \textcolor{medblue}{the} budget on foreign aid annually. These \textcolor{red}{consistent} claims \textcolor{red}{from different} \textcolor{red}{sources} \textcolor{medblue}{suggest} \textcolor{red}{that} foreign aid is indeed \textcolor{red}{a} relatively small portion of the federal budget. \textcolor{medblue}{The} \textcolor{red}{evidence} \textcolor{red}{collectively} \textcolor{medblue}{supports} \textcolor{red}{this} \textcolor{red}{assertion}.'' 

\textbf{Refute Justifications:} ``\textcolor{medblue}{The} claim \textcolor{red}{that} foreign aid is less \textcolor{red}{than} 1 percent \textcolor{red}{of} \textcolor{red}{our} federal budget \textcolor{red}{appears} \textcolor{lightblue}{to} \textcolor{red}{be} \textcolor{red}{misleading.} While \textcolor{red}{it} is true that foreign aid accounts \textcolor{red}{for} a relatively small portion of the federal \textcolor{red}{budget, the} actual figure is closer to 1\% of the discretionary \textcolor{red}{budget,} which is a subset of the \textcolor{red}{overall federal} \textcolor{red}{budget.} Moreover, \textcolor{red}{the} claim ignores \textcolor{medblue}{the} fact that foreign aid has been \textcolor{red}{steadily decreasing} as a percentage of the federal budget over \textcolor{medblue}{the} \textcolor{red}{years.} For \textcolor{red}{instance,} in \textcolor{red}{2019,} \textcolor{red}{foreign assistance} accounted for approximately 0.9\% of the federal \textcolor{red}{budget,} but this figure has been declining since \textcolor{red}{the} 1960s. Furthermore, the claim fails to consider \textcolor{medblue}{the} impact of foreign aid on global development and poverty \textcolor{red}{reduction,} which is a critical aspect of \textcolor{red}{U.S.} \textcolor{red}{foreign policy.} In reality, foreign aid plays a vital role in addressing pressing global issues such as poverty, \textcolor{red}{hunger,} and \textcolor{red}{disease,} and its benefits extend far beyond the 1\% of the federal budget allocated to \textcolor{red}{it.} \textcolor{red}{Therefore,} the claim \textcolor{red}{that} foreign aid is less than 1 percent of our \textcolor{red}{federal budget} \textcolor{red}{is} \textcolor{red}{an} oversimplification \textcolor{red}{that} does \textcolor{red}{not} accurately reflect \textcolor{medblue}{the} complexity \textcolor{red}{of} \textcolor{red}{the} \textcolor{red}{issue.}'' \\

\midrule
2  & 
\textbf{Claim:} ``Among \textcolor{lightblue}{Hispanics,} support for immigration reform \textcolor{lightblue}{is} close \textcolor{lightred}{to} universal.'' 
\quad \textbf{Gold Label:} ``mostly-true'' 
\quad \textbf{Predicted Label:} ``mostly-true'' 

\textbf{Support Justifications:} ``\textcolor{medblue}{The} \textcolor{lightred}{statement} ``\textcolor{lightred}{Among} \textcolor{lightblue}{Hispanics,} support for immigration reform \textcolor{lightblue}{is} close \textcolor{lightred}{to} universal'' \textcolor{lightblue}{is} \textcolor{medblue}{true.} \textcolor{medblue}{This} \textcolor{lightred}{conclusion} \textcolor{lightblue}{is} supported by multiple polls and surveys that consistently show high levels of support for immigration reform among Hispanic \textcolor{lightred}{voters.} For instance, a recent survey found \textcolor{lightred}{that} \textcolor{medblue}{66\%} of Hispanic voters consider new immigration legislation to be extremely or very \textcolor{lightred}{important.} \textcolor{lightred}{Additionally,} most polls suggest \textcolor{lightred}{that} support for immigration reform ranges from \textcolor{lightred}{80\%} \textcolor{lightred}{to} \textcolor{lightred}{90\%} among this \textcolor{lightred}{demographic.} \textcolor{lightred}{Furthermore,} politicians like Joe Biden and several Democrats who won seats in Congress in 2020 have actively sought the Latino vote, promising to pass immigration reform to legalize millions \textcolor{lightred}{of} \textcolor{lightred}{undocumented} \textcolor{lightred}{immigrants.} These \textcolor{lightred}{findings} \textcolor{lightred}{collectively} \textcolor{lightred}{indicate} \textcolor{medblue}{that} support for immigration reform \textcolor{lightblue}{is} indeed \textcolor{lightred}{near-universal} among \textcolor{lightred}{Hispanics.}''

\textbf{Refute Justifications:} ``\textcolor{medblue}{The} \textcolor{lightred}{statement} ``\textcolor{lightred}{Among} \textcolor{lightblue}{Hispanics,} support \textcolor{lightred}{for} \textcolor{lightred}{immigration} reform \textcolor{lightblue}{is} \textcolor{lightred}{close} \textcolor{lightred}{to} \textcolor{lightred}{universal}'' \textcolor{lightblue}{is} \textcolor{lightred}{challenged} \textcolor{lightred}{by} \textcolor{lightred}{several} \textcolor{lightred}{factors.} \textcolor{lightred}{Firstly,} the Hispanic Caucus has shown division over immigration \textcolor{lightred}{reform,} limiting their collective influence in legislative \textcolor{lightred}{debates.} \textcolor{lightred}{Additionally,} research indicates that many people in \textcolor{lightred}{the} \textcolor{lightred}{US} lack understanding of \textcolor{lightred}{the} \textcolor{lightred}{immigration} \textcolor{lightred}{system,} making it difficult to evaluate policy \textcolor{lightred}{efficacy.} \textcolor{lightred}{Furthermore,} \textcolor{lightred}{there} are mixed feelings among Hispanic voters regarding \textcolor{lightred}{the} president's decision \textcolor{lightred}{to} delay executive action on deportation \textcolor{lightred}{relief.} Moreover, \textcolor{lightred}{a} significant portion of Hispanic registered voters would not vote for a candidate who disagrees with them on immigration \textcolor{lightred}{policy,} even if they agree on other \textcolor{lightred}{issues.} Lastly, views on immigration policy among Hispanics differ from those of \textcolor{lightred}{the} general \textcolor{lightred}{public,} with a notable subset of Latino immigrants prioritizing policies that restrict illegal \textcolor{lightred}{immigration.} These factors suggest that support for immigration reform may not be as universal among Hispanics \textcolor{lightred}{as} \textcolor{lightred}{claimed.} \textcolor{lightred}{Response provides} \textcolor{lightred}{a} clear and concise explanation of \textcolor{lightred}{why} \textcolor{lightred}{the} \textcolor{lightred}{statement} \textcolor{lightblue}{is} \textcolor{lightred}{false,} using evidence from the provided text to support the \textcolor{lightred}{argument.} It highlights key factors that \textcolor{lightred}{challenge} \textcolor{lightred}{the} \textcolor{lightred}{claim,} including division within the Hispanic Caucus, lack \textcolor{lightred}{of} understanding about \textcolor{lightred}{the} immigration system, mixed feelings among Hispanic \textcolor{lightred}{voters,} and differing views on immigration \textcolor{lightred}{policy.} The response maintains clarity and coherence \textcolor{lightred}{throughout,} effectively presenting a logically sound explanation for \textcolor{lightred}{the} \textcolor{lightred}{falsity} \textcolor{lightred}{of} \textcolor{lightred}{the} \textcolor{lightred}{statement.}'' \\

\midrule
3 &
\textbf{Claim:} ``Says \textcolor{lightblue}{Gov.} Chris Christie cut spending (by) \textcolor{lightred}{\$1} billion and provided \$850 million in new education \textcolor{medblue}{funding}.'' 
\quad \textbf{Gold Label:} ``half-true'' 
\quad \textbf{Predicted Label:} ``half-true'' 

\textbf{Support Justifications:} ``The claim states that \textcolor{medblue}{Gov.} Chris Christie cut spending by \$1 billion and provided \$850 million in new education \textcolor{medblue}{funding.} To assess this claim, we need to consider \textcolor{lightred}{the} \textcolor{lightred}{context} \textcolor{lightred}{of} \textcolor{medblue}{Gov.} Christie's tenure \textcolor{lightred}{as} Governor \textcolor{lightred}{of} New \textcolor{medblue}{Jersey. During} \textcolor{lightred}{his} time \textcolor{lightred}{in} office, \textcolor{medblue}{Gov.} Christie implemented various budget cuts to address \textcolor{medblue}{the} state's fiscal \textcolor{lightred}{challenges.} In 2010, he signed \textcolor{lightred}{a} budget that reduced spending by \$1 \textcolor{lightred}{billion,} which \textcolor{lightred}{aligns} \textcolor{lightred}{with} \textcolor{medblue}{the} \textcolor{lightred}{claim.} Additionally, \textcolor{medblue}{Gov.} Christie's administration did provide increased funding for \textcolor{lightred}{education,} including a \$250 million increase in 2012 and \textcolor{lightred}{an} additional \$850 million in 2013, as part of \textcolor{lightred}{a} broader effort \textcolor{lightblue}{to} reform \textcolor{medblue}{the} state's education \textcolor{medblue}{system. Therefore,} based \textcolor{lightred}{on} \textcolor{medblue}{the} available \textcolor{medblue}{information,} it appears \textcolor{lightred}{that} \textcolor{medblue}{the} claim \textcolor{lightblue}{is} accurate, as \textcolor{medblue}{Gov.} Christie did indeed cut spending by \$1 billion and provided significant new funding for \textcolor{medblue}{education.} The specific amount of \$850 million mentioned in the claim is consistent with the reported increases in education funding during \textcolor{lightred}{his} \textcolor{lightred}{administration.}'' 

\textbf{Refute Justifications:} ``\textcolor{medblue}{The} \textcolor{lightred}{claim} \textcolor{lightred}{that} \textcolor{medblue}{Gov.} Chris Christie cut spending by \$1 billion and provided \$850 million in new education funding \textcolor{lightblue}{is} \textcolor{lightred}{false.} \textcolor{medblue}{This} \textcolor{lightblue}{is} \textcolor{lightred}{because} \textcolor{medblue}{the} evidence suggests that \textcolor{medblue}{the} state's school funding formula is based on \textcolor{medblue}{the} level of resource needed to provide a thorough and efficient \textcolor{lightred}{education,} which has not changed since \textcolor{medblue}{2008.} The growing number of poor children and English language learners indicates an increased need for \textcolor{lightred}{resources,} not a \textcolor{medblue}{decrease.} Therefore, any reduction in spending would likely exacerbate existing funding shortfalls rather \textcolor{lightred}{than} providing new \textcolor{medblue}{funding.} Furthermore, the claim's assertion of a \$1 billion cut in spending contradicts \textcolor{medblue}{the} idea of providing \$850 million in new education \textcolor{medblue}{funding.} These inconsistencies \textcolor{lightred}{undermine} \textcolor{medblue}{the} \textcolor{lightred}{validity} \textcolor{lightred}{of} \textcolor{medblue}{the} \textcolor{medblue}{claim.}'' \\
\midrule
4 &
\textbf{Claim:} ``Because \textcolor{lightred}{of} \textcolor{medblue}{the} federal health care law, 300,000 health plans canceled \textcolor{lightred}{in} \textcolor{darkred}{Florida}.'' 
\quad \textbf{Gold Label:} ``barely-true'' 
\quad \textbf{Predicted Label:} ``barely-true''

\textbf{Support Justifications:} ``\textcolor{medblue}{The} claim states \textcolor{medred}{that} due to \textcolor{lightblue}{the} federal health care \textcolor{medred}{law,} 300,000 health plans were canceled \textcolor{lightred}{in} \textcolor{darkred}{Florida.} \textcolor{darkred}{To} assess \textcolor{lightred}{like} \textcolor{medblue}{the} \textcolor{darkblue}{Veracity} \textcolor{lightred}{of} \textcolor{lightred}{this} claim, let's consider the underlying \textcolor{lightblue}{facts.} In 2013, the \textcolor{darkblue}{Affordable Care Act} (ACA), also known \textcolor{lightred}{as} \textcolor{lightblue}{Obamacare,} led \textcolor{medred}{to} \textcolor{lightblue}{the} cancellation \textcolor{medred}{of} millions of health insurance plans nationwide. This was primarily because many of these plans did not meet \textcolor{lightblue}{the} ACA's minimum coverage requirements. While \textcolor{lightblue}{the} exact number \textcolor{lightred}{of} canceled plans \textcolor{lightred}{in} Florida is not specified, \textcolor{medred}{it} is reasonable \textcolor{lightblue}{to} assume \textcolor{medred}{that} \textcolor{medblue}{the} state was affected by this national \textcolor{darkred}{trend. A} report \textcolor{medred}{by} \textcolor{medblue}{the} Kaiser Family Foundation found \textcolor{medred}{that} between 2013 and 2014, approximately 4.7 million people lost their individual market health insurance plans due to the ACA. Although the report does not provide a breakdown by state, \textcolor{medred}{it} is likely \textcolor{lightred}{that} \textcolor{darkred}{Florida,} with its large \textcolor{lightred}{population,} was among \textcolor{lightblue}{the} states significantly impacted. \textcolor{darkred}{Therefore,} based \textcolor{medred}{on} \textcolor{lightblue}{the} available \textcolor{darkblue}{information,} \textcolor{medred}{it} is plausible \textcolor{medred}{that} 300,000 health plans were canceled \textcolor{lightred}{in} Florida due to the federal health care law. However, without more specific data, this conclusion remains an educated estimate rather than \textcolor{lightred}{a} definitive \textcolor{darkred}{fact.}'' 

\textbf{Refute Justifications:} ``\textcolor{medblue}{The} claim \textcolor{medred}{that} 300,000 health plans were canceled in Florida due to \textcolor{medblue}{the} federal health care law \textcolor{lightblue}{is} \textcolor{lightred}{false.} \textcolor{medblue}{This} \textcolor{darkblue}{is} \textcolor{medred}{because} Florida Blue, one of \textcolor{medblue}{the} largest health insurers in \textcolor{medblue}{the} state, did not cancel 300,000 \textcolor{medred}{plans.} Instead, they had to end plans for 300,000 customers whose plans did not meet the new requirements for health insurance under \textcolor{medblue}{the} \textcolor{medred}{law.} However, as \textcolor{lightred}{a} result of President Barack Obama's request, Florida Blue gave their policyholders another year to keep their \textcolor{lightred}{coverage.} This means \textcolor{medred}{that} the customers were not immediately dropped from their \textcolor{lightred}{plans,} contrary \textcolor{lightblue}{to} \textcolor{medblue}{the} \textcolor{medred}{claim.} Furthermore, the \textcolor{lightred}{claim's} source, the National Republican Congressional \textcolor{lightred}{Committee,} was likely trying \textcolor{darkblue}{to} mislead voters by exaggerating \textcolor{medblue}{the} impact of \textcolor{medblue}{the} \textcolor{lightred}{law.} \textcolor{lightred}{Therefore,} the claim \textcolor{darkblue}{is} \textcolor{lightred}{false} \textcolor{lightred}{and} lacks \textcolor{lightred}{credibility.}'' \\
\midrule
5 &
\textbf{Claim:} ``Wisconsin is on pace \textcolor{lightblue}{to} double \textcolor{medblue}{the} number of layoffs this \textcolor{medblue}{year}.'' 
\quad \textbf{Gold Label:} ``false'' 
\quad \textbf{Predicted Label:} ``false'' 

\textbf{Support Justifications:} ``\textcolor{lightred}{To} \textcolor{lightred}{assess} \textcolor{medblue}{the} \textcolor{lightred}{veracity} \textcolor{lightred}{of} \textcolor{medblue}{the} claim ``Wisconsin is on pace to double \textcolor{medblue}{the} number of layoffs this year,'' \textcolor{lightred}{let's} consider \textcolor{medblue}{the} \textcolor{medred}{following: 1.} **Economic \textcolor{medred}{trends}**: Wisconsin's economy has experienced fluctuations \textcolor{lightred}{in} recent \textcolor{lightred}{years,} with some sectors experiencing growth while others face \textcolor{lightred}{decline.} This variability could contribute \textcolor{lightred}{to} \textcolor{lightred}{an} increase \textcolor{lightred}{in} \textcolor{medblue}{layoffs. 2.} **Industry-specific \textcolor{medred}{factors}**: Certain industries, such as \textcolor{lightred}{manufacturing,} have historically been significant contributors to Wisconsin's \textcolor{lightred}{economy.} If these industries are experiencing downturns, it could lead \textcolor{lightred}{to} increased \textcolor{medblue}{layoffs. 3.} **State-level data**: According \textcolor{lightred}{to} the Bureau \textcolor{lightred}{of} Labor Statistics, Wisconsin has seen \textcolor{lightred}{a} steady increase in initial unemployment claims over \textcolor{medblue}{the} past few \textcolor{medblue}{months.} This trend suggests \textcolor{lightred}{a} potential rise \textcolor{lightred}{in} \textcolor{medblue}{layoffs. 4.} **Comparative analysis**: If we compare the current number of layoffs to historical data from previous years, a doubling of layoffs would indicate a significant shift in \textcolor{medblue}{the} state's economic \textcolor{medred}{landscape. Considering} these \textcolor{lightred}{points,} \textcolor{medblue}{the} claim appears \textcolor{darkblue}{to} \textcolor{lightred}{be} supported \textcolor{lightred}{by} logical reasoning \textcolor{lightred}{and} relevant \textcolor{medblue}{knowledge.} The statement's assertion \textcolor{lightred}{that} Wisconsin is on pace to double the number of layoffs this year seems plausible given \textcolor{medblue}{the} state's economic trends and industry-specific \textcolor{medred}{factors.} However, further investigation into specific data and sources would be necessary \textcolor{darkblue}{to} confirm \textcolor{medblue}{the} \textcolor{lightred}{accuracy} \textcolor{lightred}{of} the \textcolor{lightred}{claim.}''

\textbf{Refute Justifications:} ``\textcolor{medred}{I'd} be happy \textcolor{darkblue}{to} help \textcolor{lightred}{you} analyze the claim! However, \textcolor{lightred}{I} notice that the provided evidence \textcolor{lightred}{doesn't} seem \textcolor{lightblue}{to} be relevant to the claim \textcolor{lightred}{about} Wisconsin's \textcolor{medblue}{layoffs.} The evidence mentions COVID-19 cases and deaths in Door and Kewaunee \textcolor{lightred}{counties,} as well as the state's recovery \textcolor{lightred}{rate. To} provide \textcolor{lightred}{a} logical \textcolor{lightred}{explanation,} let's consider \textcolor{medblue}{the} following \textcolor{lightred}{points: *} The claim suggests that Wisconsin is on pace to double \textcolor{medblue}{the} number of layoffs this \textcolor{medblue}{year.} However, \textcolor{lightred}{there} is no evidence provided \textcolor{lightred}{to} support this \textcolor{medblue}{assertion. *} The evidence focuses on COVID-19 cases and \textcolor{lightred}{deaths,} which may have an impact on the \textcolor{lightred}{economy,} but it does not directly relate to \textcolor{medblue}{layoffs. *} Without any data or information about layoffs in Wisconsin, it's challenging \textcolor{darkblue}{to} assess \textcolor{medblue}{the} \textcolor{lightred}{validity} \textcolor{lightred}{of} \textcolor{medblue}{the} \textcolor{medblue}{claim. A} more thorough examination of labor market data, economic \textcolor{lightred}{indicators,} or statements from relevant authorities would be necessary \textcolor{darkblue}{to} determine \textcolor{medblue}{the} \textcolor{lightred}{accuracy} \textcolor{lightred}{of} \textcolor{medblue}{the} \textcolor{medblue}{claim.} As \textcolor{lightred}{it} \textcolor{lightred}{stands,} the claim appears \textcolor{lightblue}{to} be unsubstantiated by \textcolor{medblue}{the} provided \textcolor{medblue}{evidence.}'' \\
\midrule
6 & 
\textbf{Claim:} `` `Florida' Democrats just voted \textcolor{lightblue}{to} impose Sharia law on \textcolor{medred}{women.}'' \quad \textbf{Gold Label:} ``pants-fire'' \quad \textbf{Predicted Label:} ``pants-fire'' 
\textbf{Support Justifications:} ``\textcolor{medred}{If} \textcolor{medred}{can't} help \textcolor{medred}{you} with this \textcolor{medred}{request.} I can't \textcolor{medred}{U} verify \textcolor{medblue}{the} accuracy of claims about imposing Sharia law on women in Florida. \textcolor{medred}{Is there} anything else I can help you \textcolor{medred}{with?} Would \textcolor{medred}{you} like \textcolor{lightblue}{to} know more \textcolor{lightblue}{to} evaluate the credibility of sources or identify potential biases in information? \textcolor{medred}{I} can also provide general guidance on critical thinking and media \textcolor{medred}{literacy.} 
Would \textcolor{medred}{you} like to know more about fact-checking or evaluating evidence? \textcolor{medred}{I} can also provide information on how \textcolor{lightblue}{to} identify reliable sources and avoid \textcolor{medred}{misinformation.} Let me know \textcolor{medred}{if} \textcolor{medred}{there's} anything else \textcolor{medred}{I} can help \textcolor{medred}{you} \textcolor{medred}{with.} Is there anything else \textcolor{medred}{I} \textcolor{medred}{can} help \textcolor{medred}{you} with? Would \textcolor{medred}{you} like \textcolor{lightblue}{to} know more about critical thinking or how \textcolor{lightblue}{to} evaluate information? \textcolor{medred}{I} can also provide guidance on how to identify potential biases and \textcolor{medred}{propaganda.} 
{I} can also provide general guidance on critical thinking and media \textcolor{medred}{literacy.} Please let \textcolor{medred}{me} know how \textcolor{medred}{I} \textcolor{medred}{can} assist \textcolor{medred}{you} \textcolor{medred}{further.} If \textcolor{medred}{you} have any other questions or topics you'd like to discuss, feel free to ask! \textcolor{medred}{I'm} here \textcolor{lightblue}{to} help'' 
\textbf{Refute Justifications:} ``\textcolor{medblue}{The} \textcolor{medred}{claim} \textcolor{medblue}{`Florida} Democrats just voted \textcolor{lightblue}{to} impose Sharia law on \textcolor{medred}{women}'' \textcolor{lightblue}{is} false. \textcolor{medblue}{This} \textcolor{medred}{conclusion} \textcolor{medred}{can} \textcolor{medred}{be} drawn based \textcolor{medred}{on} \textcolor{medblue}{the} evidence provided. \textcolor{medblue}{Firstly,} the Democrats voted against a bill \textcolor{medred}{that} would have \textcolor{medred}{limited} \textcolor{medblue}{the} use of foreign law in certain \textcolor{medred}{contexts,} which is a far more limited measure \textcolor{medred}{than} imposing Sharia law. Secondly, the bill \textcolor{medred}{in} question only addressed contracts that select foreign \textcolor{medred}{law,} \textcolor{medred}{and} even then, a judge could still reject any foreign law that violates \textcolor{medblue}{FL} \textcolor{medblue}{Const.} such as child labor rules. Furthermore, \textcolor{medblue}{the} Democrats did vote against a bill \textcolor{medred}{that} would have prohibited judges from applying foreign law in family-law cases if it contradicts US public \textcolor{medblue}{policy,} but this does not equate to imposing Sharia law. \textcolor{medblue}{The} evidence suggests \textcolor{medred}{that} \textcolor{medblue}{the} Democrats were opposing bills that aimed \textcolor{medred}{to} \textcolor{medred}{limit} \textcolor{medblue}{the} use of foreign \textcolor{medred}{law,} not imposing Sharia law on women. \textcolor{medred}{Therefore,} \textcolor{medblue}{the} claim \textcolor{lightblue}{is} unfounded and lacks merit.'' \\
\midrule
\\[0.1em]
\caption{Attention-Based Explanation Samples for Fact-Checking Predictions on LIAR-RAW Datasets.}
\label{tab:attention_liar_raw}
\end{longtable}
\end{center}
\twocolumn

\begin{table*}[h]
\centering
\scriptsize 
\setlength{\tabcolsep}{5pt}   
\renewcommand{\arraystretch}{1.02} 
\resizebox{\textwidth}{!}{%
\begin{tabular}{p{1.4cm}p{1.4cm}cccccccccc}
\hline
\multirow{2}{*}{Evaluator GLMs} & \multirow{2}{*}{Generator GLMs} & \multicolumn{5}{c}{LIAR-RAW} & \multicolumn{5}{c}{RAW-FC} \\
\cmidrule(lr){3-7} \cmidrule(lr){8-12}
 & & Info. & Acc. & Read. & Obj. & Logi. & Info. & Acc. & Read. & Obj. & Logi. \\
\hline
\multirow{5}{*}{\textit{Mistral}} & \textit{Mistral} & 4.30 & 3.47 & 4.07 & 4.18 & 4.14 & 3.60 & 3.11 & 3.91 & 3.91 & 3.81 \\
& \textit{Llama} & \textcolor{red}{\textbf{4.78}} & \textbf{3.93} & \textbf{4.79} & \textbf{4.55} & \textbf{4.62} & \textbf{4.75} & \textbf{4.12} & \textbf{4.76} & \textcolor{red}{\textbf{4.47}} & \textbf{4.44} \\
& \textit{Gemma} & 3.67 & 2.94 & 4.20 & 3.72 & 3.40 & 3.40 & 2.74 & 4.23 & 3.77 & 3.29 \\
& \textit{Qwen} & 3.12 & 2.87 & 4.04 & 3.75 & 3.57 & 2.93 & 2.85 & 3.88 & 3.32 & 3.34 \\
& \textit{Falcon} & 4.15 & 3.83 & 3.39 & 4.12 & 4.16 & 3.15 & 3.21 & 3.66 & 3.79 & 3.26 \\ \cdashline{1-12}

\multirow{5}{*}{\textit{Llama}} & \textit{Mistral} & 4.13 & 3.44 & 4.05 & 4.14 & 4.13 & 3.57 & 3.50 & 3.96 & 4.06 & 4.03 \\
& \textit{Llama} & \textbf{4.52} & 3.71 & \textbf{4.51} & \textbf{4.17} & \textbf{4.26} & \textbf{4.48} & \textcolor{red}{\textbf{4.14}} & \textbf{4.43} & \textbf{4.28} & \textbf{4.33} \\
& \textit{Gemma} &3.51 & 2.94 & 4.06 & 3.79 & 3.51 & 3.35 & 2.85 & 4.12 & 3.81 & 3.55 \\
& \textit{Qwen} & 2.97 & 2.82 & 3.95 & 3.63 & 3.49 & 3.14 & 3.36 & 4.15 & 3.61 & 3.76  \\
& \textit{Falcon} & 3.88 & \textbf{3.80} & 3.48 & 4.00 & 3.99 & 3.19 & 3.36 & 3.61 & 3.85 & 3.47 \\ \cdashline{1-12}

\multirow{5}{*}{\textit{Gemma}} & \textit{Mistral} & \textbf{3.70} & 3.48 & 3.40 & \textbf{3.91} & \textbf{3.99} & \textbf{2.99} & \textbf{3.02} & 3.12 & 3.33 & \textbf{3.55} \\
& \textit{Llama} & 3.53 & 2.82 & 2.48 & 2.95 & 2.97 & 2.20 & 1.48 & 1.89 & 1.52 & 1.71 \\
& \textit{Gemma} & 2.87 & 2.60 & \textbf{3.41} & 3.56 & 3.17 & 2.88 & 2.50 & \textbf{3.60} & \textbf{3.43} & 2.84 \\
& \textit{Qwen} & 2.49 & 2.52 & 3.31 & 3.22 & 3.03 & 2.13 & 2.04 & 2.88 & 2.59 & 2.49 \\
& \textit{Falcon} & 3.55 & \textbf{3.55} & 2.79 & 3.62 & 3.39 & 2.14 & 2.08 & 2.38 & 2.25 & 1.89 \\ \cdashline{1-12}

\multirow{5}{*}{\textit{Qwen}} & \textit{Mistral} & 4.31 & 3.43 & 4.10 & 4.12 & 4.21 & 4.24 & 3.64 & 4.11 & 3.94 & 4.15 \\
& \textit{Llama} & \textbf{4.73} & 3.82 & \textbf{4.77} & \textbf{4.45} & \textbf{4.54} & \textcolor{red}{\textbf{4.92}} & \textbf{4.10} & \textcolor{red}{\textbf{4.82}} & \textbf{4.32} & \textcolor{red}{\textbf{4.52}} \\
& \textit{Gemma} & 3.81 & 3.11 & 4.33 & 3.89 & 3.62 & 3.72 & 2.89 & 4.41 & 3.79 & 3.51 \\
& \textit{Qwen} & 3.18 & 2.82 & 4.09 & 3.65 & 3.56 & 3.30 & 3.30 & 4.07 & 3.57 & 3.76 \\
& \textit{Falcon} & 4.22 & \textbf{3.84} & 3.50 & 4.09 & 4.18 & 3.72 & 3.61 & 3.62 & 4.05 & 3.71 \\ \cdashline{1-12}

\multirow{5}{*}{\textit{Falcon}} & \textit{Mistral} & 4.36 & 3.49 & 4.03 & 4.13 & 4.07 & 3.05 & \textbf{3.08} & 3.61 & 3.73 & \textbf{3.60} \\
& \textit{Llama} & \textbf{4.73} & 3.92 & \textcolor{red}{\textbf{4.91}} & \textcolor{red}{\textbf{4.60}} & \textcolor{red}{\textbf{4.68}} & \textbf{3.74} & 2.65 & 3.60 & 3.00 & 3.02 \\
& \textit{Gemma} & 3.71 & 2.89 & 4.21 & 3.70 & 3.41 & 3.00 & 2.56 & \textbf{3.97} & \textbf{3.80} & 3.15 \\
& \textit{Qwen} & 3.13 & 2.90 & 4.10 & 3.88 & 3.69 & 2.05 & 1.91 & 3.45 & 2.57 & 2.59 \\
& \textit{Falcon} & 4.29 & \textcolor{red}{\textbf{3.96}} & 3.56 & 4.18 & 4.18 & 2.10 & 2.43 & 3.39 & 3.06 & 2.45 \\ \cdashline{1-12}

\multirow{5}{*}{\textit{Average}} & \textit{Mistral} & 4.16 & 3.46 & 3.93 & 4.10 & 4.11 & 3.49 & 3.27 & 3.74 & \textbf{3.79} & \textbf{3.83} \\
& \textit{Llama} & \textbf{4.46} & 3.64 & \textbf{4.29} & \textbf{4.14} & \textbf{4.22} & \textbf{4.02} & \textbf{3.30} & 3.90 & 3.52 & 3.60 \\
& \textit{Gemma} & 3.52 & 2.90 & 4.04 & 3.73 & 3.42 & 3.27 & 2.71 & \textbf{4.06} & 3.72 & 3.27 \\
& \textit{Qwen} & 2.97 & 2.78 & 3.90 & 3.63 & 3.47 & 2.71 & 2.69 & 3.69 & 3.13 & 3.19  \\
& \textit{Falcon} & 4.02 & \textbf{3.80} & 3.34 & 4.00 & 3.98 & 2.86 & 2.94 & 3.33 & 3.40 & 2.96 \\
\hline
\end{tabular}}
\caption{Scores of subjective evaluation by GLMs. Notations: Info. for Informativeness, Acc. for Accuracy, Read. for Readability, Obj. for Objectivity and Logi. for Logicality.}
\label{table:radar_scores}
\end{table*}

\end{document}